%% file: top.tex
\documentclass[10pt,twocolumn,letterpaper]{article}

\usepackage{float}
\usepackage{cvpr}
\usepackage{times}
\usepackage{epsfig}
\usepackage{graphicx}
\usepackage{amsmath}
\usepackage{amssymb}
\usepackage{cite} %
\usepackage{csquotes} %

\usepackage{multirow} %
\usepackage{listings}
\usepackage{graphicx}
\usepackage{enumitem} %
\usepackage{physics}
\usepackage{color}
\usepackage{tabularx}
\usepackage{booktabs} %
\usepackage{wrapfig}
\usepackage{pifont}
\usepackage[font=small,bf]{caption}
\usepackage{algorithm}
\usepackage{algpseudocode}
\usepackage{algorithmicx}
\usepackage{xparse}
\usepackage{microtype}
\usepackage{longtable}

\makeatletter
\NewDocumentCommand{\LeftComment}{s m}{%
\textit{\Statex \IfBooleanF{#1}{\hspace*{\ALG@thistlm}}#2}}
\makeatother

\makeatletter
\renewcommand{\paragraph}{%
  \@startsection{paragraph}{4}%
  {\z@}{1.25ex \@plus 1ex \@minus .2ex}{-1em}%
  {\normalfont\normalsize\bfseries}%
}
\makeatother

\newcommand{\cmark}{\ding{51}}%

\graphicspath{{./imgs}}

\renewcommand{\vec}[1]{\mathbf{#1}}

\newcommand{\R}{\mathbb{R}}

\newfloat{lstfloat}{htbp}{lop}
\floatname{lstfloat}{Listing}

\definecolor{julieta_colour}{RGB}{117,112,178} %
\definecolor{jashan_colour}{RGB}{27,158,119}   %
\definecolor{raquel_colour}{RGB}{217,95,2}     %
\definecolor{wenyuan_colour}{RGB}{231,41,138}  %
\definecolor{andrei_colour}{RGB}{180,180,32}

\definecolor{mygreen}{RGB}{0, 128, 0}
\definecolor{mygrey}{gray}{.98}
\definecolor{darkgrey}{gray}{.60}

\definecolor{citecolor}{RGB}{34,139,34}
\usepackage[pagebackref=true,breaklinks=true,letterpaper=true,colorlinks, citecolor=citecolor,bookmarks=false]{hyperref}

\DeclareFixedFont{\ttb}{T1}{txtt}{bx}{n}{8} %
\DeclareFixedFont{\ttm}{T1}{txtt}{m}{n}{8}  %

\definecolor{deepblue}{rgb}{0,0,0.5}
\definecolor{deepred}{rgb}{0.6,0,0}
\definecolor{deepgreen}{rgb}{0,0.5,0}

\DeclareMathOperator*{\argmin}{arg\,min}

\usepackage{listings}

\newcommand\pythonstyle{\lstset{
language=Python,
basicstyle=\tiny\ttm,
otherkeywords={self},             %
keywordstyle=\ttb\color{deepblue},
emph={CompressedLinear,__init__},          %
emphstyle=\ttb\color{deepred},    %
stringstyle=\color{deepgreen},
frame=tb,                         %
showstringspaces=false            %
}}

\newcommand{\mytilde}{\raise.17ex\hbox{$\scriptstyle\sim$}}
\newcommand{\Bt}{\vec{B}_t}
\newcommand{\Ct}{\mathcal{C}_t}
\newcommand{\wPtij}{\vec{w}^{\vec{P}_t}_{i,j}}
\newcommand{\PWt}{\vec{P}_t\vec{W}_t}

\lstnewenvironment{python}[1][]
{
\pythonstyle
\lstset{#1}
}
{}

\newcommand\pythoninline[1]{{\pythonstyle\lstinline!#1!}}

\cvprfinalcopy %

\newif\ifarxiv
\arxivtrue

\def\cvprPaperID{4812} %

\ifcvprfinal\fi
\begin{document}

\title{\vspace{-1em}Permute, Quantize, and Fine-tune: Efficient Compression of Neural Networks}

\author{
   Julieta Martinez$^{*,1}$ \quad Jashan Shewakramani$^{*,1,2}$ \quad Ting Wei Liu$^{*,1,2}$ \\
   Ioan Andrei B{\^a}rsan$^{1, 3}$ \quad Wenyuan Zeng$^{1, 3}$ \quad Raquel Urtasun$^{1, 3}$ \\ \\
   $^{1}$Uber Advanced Technologies Group  \quad \quad
   $^{2}$University of Waterloo \quad \quad
   $^{3}$University of Toronto\\
  \small\texttt{\{julieta,jashan,tingwei.liu,andreib,wenyuan,urtasun\}@uber.com}
}

\maketitle

\ifarxiv
  \thispagestyle{plain}
  \pagestyle{plain}
\else
  \thispagestyle{empty}
\fi

\begin{abstract}
   \input{sections/abstract}
\end{abstract}
\input{sections/introduction}

\input{sections/related}
\input{sections/method}

\input{sections/experiments}
\input{sections/conclusions}

{\small
\bibliographystyle{ieee_fullname}
\bibliography{egbib}
}

\ifarxiv
\onecolumn
\section*{Appendices}

\input{supp}

\fi

\end{document}

%% file: sections/abstract.tex
Compressing large neural networks is an important step for their deployment in resource-constrained computational platforms.
In this context, vector quantization is an appealing framework that expresses multiple parameters using a single code, and  has recently achieved state-of-the-art network compression on a range of core vision and natural language processing tasks.
Key to the success of vector quantization  is deciding which parameter groups should be compressed together.
Previous work has relied on heuristics that group the spatial dimension of individual convolutional filters,
but a general solution remains unaddressed.
This is desirable for pointwise convolutions (which dominate modern architectures),
linear layers (which have no notion of spatial dimension),
and convolutions (when more than one filter is compressed to the same codeword).
In this paper we make the observation that the weights of two adjacent layers can be permuted while expressing the same function.
We then establish a connection to rate-distortion theory and search for permutations that result in networks that are easier to compress.
Finally, we rely on an annealed quantization algorithm to better compress the network and achieve higher final accuracy.
We show results on image classification, object detection, and segmentation,
reducing the gap with the uncompressed model by 40 to 70\% \wrt the current state of the art.
\ifcvprfinal{All our experiments can be reproduced using the code at \url{https://github.com/uber-research/permute-quantize-finetune}.
}\else{We will release code to reproduce all our experiments.}\fi

%% file: sections/introduction.tex
\section{Introduction}

State-of-the-art approaches to many computer vision tasks are currently based on deep neural networks.
These networks often have large memory and computational requirements,
limiting the range of hardware platforms on which they can operate.
This poses a challenge for applications such as virtual reality and robotics,
which naturally rely on mobile and low-power computational platforms for large-scale deployment.
At the same time, these networks are often overparameterized~\cite{denil2013predicting},
which implies that it is possible to compress them -- thereby reducing their memory and computation demands --
without much loss in accuracy.

Scalar quantization is a popular approach to network compression where
each network parameter is compressed individually, thereby limiting the achievable compression rates. %
To address this limitation, a recent line of work has focused on
vector quantization (VQ)~\cite{wu2016quantized,gong2014compressing,son2018clustering},
which compresses multiple parameters into a single code. 
Conspicuously, these approaches have recently achieved state-of-the-art compression-to-accuracy ratios on
core computer vision and natural language processing tasks~\cite{bitgoesdown, fan2020training}.

A key advantage of VQ is that it can naturally exploit redundancies among groups of network parameters,
for example, by grouping the spatial dimensions of convolutional filters in a single vector to achieve high compression rates.
However, finding which network parameters should be compressed jointly can be challenging;
for instance, there is no notion of spatial dimension in fully connected layers,
and it is not clear how vectors should be formed when the vector size is larger than a single convolutional filter --
which is always true for pointwise convolutions.
Current approaches either employ clustering (\eg, $k$-means)
using the order of the weights as
obtained by the network~\cite{gong2014compressing, wu2016quantized, son2018clustering}, which is suboptimal,
or search for groups of parameters that, when compressed jointly,
minimize the reconstruction error of the network activations~\cite{wu2016quantized, bitgoesdown, fan2020training}, which is hard to optimize.

In this paper,
we formalize the notion of redundancy among parameter groups using concepts from rate-distortion theory,
and leverage this analysis to search for permutations of the network weights that yield functionally equivalent,
yet easier-to-quantize networks.
The result is Permute, Quantize, and Fine-tune (PQF),
an efficient algorithm that first searches for permutations, codes and codebooks 
that minimize the reconstruction error of the network weights,
and then uses gradient-based optimization to recover the accuracy of the uncompressed network.
Our main contributions can be summarized as follows:
\begin{enumerate}[topsep=0.5em, partopsep=0em, itemsep=0em]
\item We study the invariance of neural networks under permutation of their weights, focusing 
on constraints induced by the network topology. We then formulate a \emph{permutation optimization} problem
to find functionally equivalent networks that are easier to quantize.
Our result focuses on improving a quantization lower bound of the weights; therefore
\item We use an efficient \emph{annealed quantization algorithm} that 
reduces quantization error and 
leads to higher accuracy of the compressed networks. Finally,
\item We show that the reconstruction error of the network parameters is \emph{inversely correlated}
with the final network accuracy after gradient-based fine-tuning.
\end{enumerate}
Put together, the above contributions define a novel method that produces 
state-of-the-art results in terms of model size vs.\ accuracy.
We benchmark our method by compressing popular architectures for image classification, and
object detection \& segmentation,
showcasing the wide applicability of our approach.
Our results show a 40-60\% relative error reduction on Imagenet object classification over the current state-of-the-art
when compressing a ResNet-50~\cite{resnet} down to about 3~MB (\mytilde$31\times$ compression).
We also demonstrate a relative 60\% (resp. 70\%) error reduction in object detection (resp. mask segmentation) on COCO over previous work,
by compressing a Mask-RCNN architecture down to about 6.6 MB (\mytilde$26\times$ compression).

%% file: sections/related.tex
\section{Related Work}

There is a vast literature on compressing neural networks.
Efforts in this area can broadly be divided into pruning, low-rank approximations, and quantization.

\paragraph{Weight pruning:}
In its simplest form, weight pruning can be achieved by removing small weights~\cite{han2015learning,guo2016dynamic},
or approximating the importance of each parameter using second-order terms~\cite{lecun1990optimal, hassibi1993second, dong2017learning}.
More sophisticated approaches use meta-learning to obtain pruning policies that generalize to multiple models~\cite{he2018amc},
or use regularization terms during training to reduce parameter count~\cite{liu2017learning}.
Most of  these methods prune individual weights,
and result in sparse networks that are difficult to accelerate on commonly available hardware.
To address these issues, another line of work aims to remove unimportant channels,
producing networks that are easier to accelerate in practice~\cite{he2017channel, luo2017thinet, li2016pruning}.

\paragraph{Low-rank approximations:}
These methods can achieve acceleration by design~\cite{denton2014exploiting,jaderberg2014speeding, lebedev2014speeding, novikov2015tensorizing},
as they typically factorize the original weight matrix into several smaller matrices.
As a result, the original computationally-heavy forward pass can be replaced by a multiplication of several smaller vectors and matrices.

\paragraph{Scalar quantization:}
These techniques constrain the number of bits that each parameter may take,
in the extreme case using binary~\cite{courbariaux2016binarized, rastegari2016xnor, martinez2020training, wang2019learning} or ternary~\cite{zhu2016trained} values.
8-bit quantization methods have proven robust and efficient,
which has motivated their native support by popular deep learning libraries such as
PyTorch\footnote{\url{pytorch.org/docs/stable/quantization.html}} and 
Tensorflow Lite\footnote{\url{tensorflow.org/lite/performance/post_training_quantization}},
with acceleration often targeting CPUs.
We refer the reader to the survey by~\cite{neill2020overview} for a recent comprehensive overview of the subject.
In this context, reducing each parameter to a single bit yields a theoretical compression ratio of $32\times$
(although, in practice, fully-connected and batch norm layers are not quantized~\cite{martinez2020training}).
To obtain higher compression ratios, researchers have turned to vector quantization.

\paragraph{Vector quantization (VQ):}
VQ of neural networks was pioneered by Gong~\etal~\cite{gong2014compressing}, who investigated scalar, vector,
and product quantization~\cite{productquantization} (PQ) of fully-connected (FC) layers,
which were the most memory-demanding layers of convolutional neural networks (CNNs) at the time.
Wu~\etal~\cite{wu2016quantized} used PQ to compress both FC and convolutional layers of CNNs;
they noticed that minimizing the quantization error of the network parameters
produces much worse results than minimizing the error of the activations,
so they sequentially quantized the layers to minimize error accumulation.
However, neither Gong~\etal~\cite{gong2014compressing} nor Wu~\etal~\cite{wu2016quantized}, explored end-to-end training,
which is necessary to recover the network accuracy as the compression ratio increases.

Son~\etal~\cite{son2018clustering} clustered 3$\times$3 convolutions using vector quantization, and fine-tuned the centroids
via gradient descent using additional bits to encode filter rotation, resulting in very compact codebooks.
However, they did not explore the compression of FC layers nor pointwise convolutions (which dominate modern architectures),
and did not explore the relationship of quantization error to accuracy.

Stock~\etal~\cite{bitgoesdown} use PQ to compress convolutional and FC layers
using a clustering technique designed to minimize the reconstruction error of the layer outputs
(which is computationally expensive),
followed by end-to-end training of the cluster centroids via distillation.
However, their approach does not optimize the grouping of the network parameters for quantization,
which we find to be crucial to obtain good compression.
Chen~\etal~\cite{chentowards} improve upon the results of~\cite{bitgoesdown} by
minimizing the reconstruction error of the parameters and the task loss jointly;
however, their method also uses more fine-tuning epochs, so a direct comparison is hard.

Different from previous approaches, our method exploits the invariance of
neural networks under permutation of their weights for the purpose of vector compression.
Based on this observation, we draw connections to rate distortion theory,
and use an efficient permutation optimization algorithm that makes the network easier to quantize.
We also use an annealed clustering algorithm to further reduce quantization error,
and show that there is a direct correlation between the quantization error of a
network weights and its final accuracy after fine-tuning.
These contributions result in an efficient method that largely outperforms its competitors on a wide range of applications.

%% file: sections/method.tex
\section{Learning to Compress a Neural Network}

In this paper we compress a neural network by compressing the weights of its layers.
Specifically, instead of storing the weight matrix $ \vec{W} $ of a layer explicitly,
we learn an encoding $ \mathcal{B} (
 \vec{W} )$ that takes considerably less memory. 
Intuitively, we can decode $ \mathcal{B} $ to a matrix $ \vec{\widehat{W}} $
that is ``close" to $ \vec{W} $, and use $ \vec{\widehat{W}} $ as the weight matrix for the layer.
The idea is  that  if $ \vec{\widehat{W}} $ is similar to $ \vec{W} $, the activations
of the layer should also be similar. Note that  the encoding will be different for each of the layers. 

\subsection{Designing the Encoding}

For a desired compression rate, we design the encoding $\mathcal{B}$ to consist of
a codebook $\mathcal{C}$, a set of codes $\vec{B}$, and a permutation matrix $\vec{P}$.
The permutation matrix preprocesses the weights so that they are easier to compress without affecting the input-output mapping of the network,
while the codes and codebook attempt to express the permuted weights as accurately as possible using limited memory.

\paragraph{Codes and codebook:}
Let $ \vec{W} \in \R^{m \times n} $ denote the weight matrix of a fully-connected (FC) layer,
with  $m$  the input size of the layer, and $n$ the size of its output.
We split each column of $\vec{W}$ into column subvectors $\vec{w}_{i,j} \in \R^{d \times 1}$, which are then compressed individually:
\begin{equation}
  \vec{W} =
  \begin{bmatrix}
      \vec{w}_{1,1} & \vec{w}_{1,2} & \cdots & \vec{w}_{1,n} \\
      \vec{w}_{2,1} & \vec{w}_{2,2} & \cdots & \vec{w}_{2,n} \\
      \vdots & \vdots & \ddots & \vdots \\
      \vec{w}_{\hat{m},1} & \vec{w}_{\hat{m},2} & \cdots & \vec{w}_{\hat{m},n}
  \end{bmatrix},
\end{equation}
where $\hat{m} = m/d$, and $\hat{m}\cdot n$ is the total number of subvectors. Intuitively, larger $d$ results in fewer subvectors and thus higher compression rates.
The set $ \{ \vec{w}_{i,j} \} $ is thus a collection of $d$-dimensional blocks that can be used to construct $ \vec{W} $.

Instead of storing all these subvectors, we approximate them by a smaller set 
$\mathcal{C} = \{ \vec{c}(1), \ldots, \vec{c}(k) \} \subseteq \R^{d \times 1}$, 
which we call the \emph{codebook}
for the layer.
We refer to the elements of $ \mathcal{C} $ as \emph{centroids}.
Let $ b_{i,j} \in \{1, \ldots, k\} $  be the index of the element in $ \mathcal{C} $
that is closest to $ \vec{w}_{i,j} $ in Euclidean space:
\begin{equation}
  b_{i,j} = \argmin_{t} \norm{\vec{w}_{i,j} - \vec{c}(t)}_2^2,
\end{equation}
The \emph{codes} $\vec{B} = \{b_{i,j}\}$ are the indices of the codes in the codebook that best reconstruct every subvector $\{ \vec{w}_{i,j} \}$.
The approximation $ \vec{\widehat{W}} $ of $ \vec{W} $ is thus the matrix obtained by replacing each subvector $ \vec{w}_{i,j} $ with $ \vec{c}(b_{i,j}) $:
\begin{equation}
  \vec{\widehat{W}} =
  \begin{bmatrix}
      \vec{c}(b_{1,1}) & \vec{c}(b_{1,2}) & \cdots & \vec{c}(b_{1,n}) \\
      \vec{c}(b_{2,1}) & \vec{c}(b_{2,2}) & \cdots & \vec{c}(b_{2,n}) \\
      \vdots & \vdots & \ddots & \vdots \\
      \vec{c}(b_{\hat{m},1}) & \vec{c}(b_{\hat{m},2}) & \cdots & \vec{c}(b_{\hat{m},n})
  \end{bmatrix}.
\end{equation}
We refer to the process of expressing the weight matrix in terms of codes and a codebook as \emph{quantization}.

\begin{figure*}[!bt]
	\centering
  \vspace{-1em}
  \includegraphics[width=0.95\linewidth,trim=5mm 140mm 4mm 5mm,clip=true]{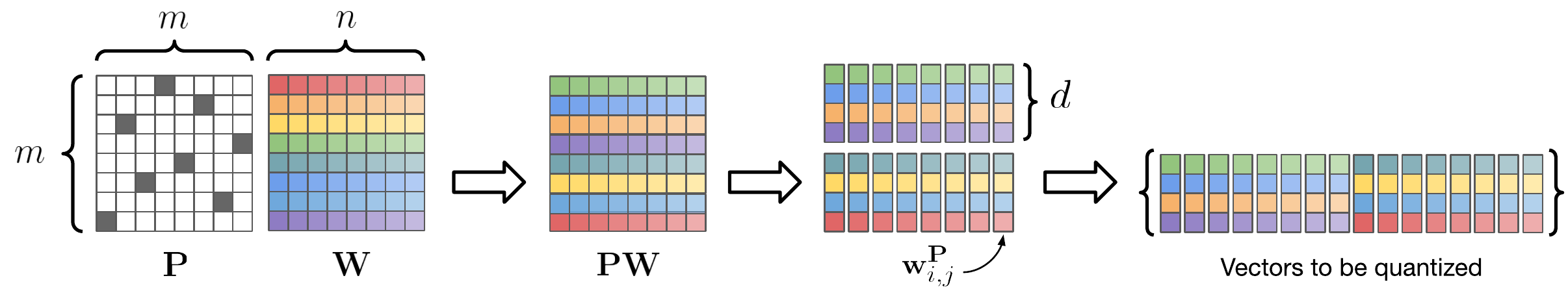}
	\caption{
    \textbf{Permutation optimization of a fully-connected layer.} Our goal is to find a permutation $\vec{P}$
    of the weights $\vec{W}$ such that the resulting subvectors are easier to compress.
  }
  \label{fig:permutation}
  \vspace{-2mm}
\end{figure*}

\paragraph{Permutation:}
The effectiveness of a given set of codes and codebooks depends on their ability to represent the original weight matrix $\vec{W}$ accurately.
Intuitively, this is easier to achieve if the subvectors $\vec{w}_{i,j}$ are similar to one another.
Therefore, it is natural to consider transformations of $\vec{W}$ that make the resulting subvectors easier to compress.

A feedforward network can be thought of as a directed acyclic graph (DAG), where nodes represent layers and edges represent the 
flow of information in the network.
We refer to the starting node of an edge as a \emph{parent} layer, and to the end node as a \emph{child} layer.
We note that the network is invariant under permutation of its weights,
as long as the same permutation is applied to the output dimension for parent layers and the input dimension for children layers.
Here, \emph{our key insight is that we can search for permutations that make the network easier to quantize}.

Formally, consider a network comprised of two layers:
\begin{align}
f(\vec{x}) &= \phi(\vec{xW}_2),\, \vec{W}_2 \in \R^{m \times n}  \\
g(\vec{x}) &= \phi(\vec{xW}_1),\, \vec{W}_1 \in \R^{p \times m}
\end{align}
where $\phi$ represents a non-linear activation function.
The network can be described as the function
\begin{align}
  f \circ g(\vec{x}) = f(g(\vec{x})) &= \phi(g(\vec{x})\vec{W}_2),
\end{align}
where $\vec{x} \in \R^{1\times p}$ is the input to the network.
Furthermore, from a topological point of view, $g$ is the parent of $f$.

Given a permutation $\pi$ of $m$ elements $\pi: \{1, \dots, m\} \rightarrow \{1, \dots, m\}$,
  we denote $\vec{P}$ as the \emph{permutation matrix} that results from reordering the rows of the $m \times m$ identity matrix according to $\pi$.
  Left-multiplying $\vec{P}$ with $\vec{X}$ has the effect of reordering the rows of $\vec{X}$ according to $\pi$.

Let $f_{\vec{P}_2}$ be the layer that results from applying the permutation matrix $\vec{P}_2$
to the input dimension of the weights of $f$:
\begin{equation}
  f_{\vec{P}_2}(\vec{x}) = \phi(\vec{xP}_2\vec{W}_2).
\end{equation}

Analogously, let $g^{\vec{P}_2}$
be the layer that results from applying the permutation $\vec{P}_2$
to the output dimension of the weights of $g$:
\begin{equation}
  g^{\vec{P}_2}(\vec{x}) = \phi(\vec{x} (\vec{P}_2\vec{W}^{\top}_1)^\top) = \phi(\vec{x} \vec{W}_1\vec{P}_2^\top).
\end{equation}

Importantly, so long as $\phi$ is an element-wise operator, $g^{\vec{P}_2}$ produces the same output as $g$, only permuted:
\begin{align}
  g^{\vec{P}_2}(\vec{x}) = g(\vec{x}) \vec{P}_2^\top,
\end{align}
then we have
\begin{align}
f_{\vec{P}_2} \circ g^{\vec{P}_2}(\vec{x}) &=  f_{\vec{P}_2}(g^{\vec{P}_2}(\vec{x}))\\
                                           &= f_{\vec{P}_2}(g(\vec{x}) \vec{P}_2^\top) \\
                                           &= \phi(g(\vec{x}) \vec{P}_2^\top \vec{P}_2\vec{W}_2) \\
                                           &= \phi(g(\vec{x}) \vec{W}_2) \\
                                           &= f(g(\vec{x})) \\
                                           &= f \circ g(\vec{x}), \,\,\, \forall\, \vec{P}_2, \vec{x}.
\end{align}
This \emph{functional equivalence} has previously been used to characterize the optimization landscape of neural networks~\cite{chen1993geometry, orhan2017skip}.
In contrast, here we focus on quantizing the permuted weight $\vec{P}_2\vec{W}_2$,
and denote its subvectors as $\{ \vec{w}^{\vec{P}_2}_{i,j} \}$.
We depict the process of applying a permutation and obtaining new subvectors in Figure~\ref{fig:permutation}.

\paragraph{Extension to convolutional layers:}
The encoding of convolutional layers is closely related to that of fully-connected layers.
Let $\vec{W} \in \R^{C_{\mathrm{in}} \times C_{\mathrm{out}} \times K \times K }$ denote the weights 
of a convolutional layer with $ C_{\mathrm{in}} $ input channels,
$C_{\mathrm{out}} $ output channels, and a kernel size of $K \times K$.
The idea is to reshape $\vec{W}$ into a 2d matrix $\vec{W_r}$ of size $C_{\mathrm{in}} K^2 \times C_{\mathrm{out}}$,
and then apply the same encoding method that we use with fully-connected layers.
The result is an approximation $\vec{\widehat{W}_r}$ to $\vec{W_r}$.
We then apply the inverse of the reshaping operation on $\vec{\widehat{W}_r}$ to get our approximation to $\vec{W}$.

When $K > 1$, we set the codeword size $d$ to a multiple of $K^2$ and
limit the permutation matrix $\vec{P}$ to have a block structure such that the spatial dimensions of filters are quantized together.
For pointwise convolutions (\ie, $K = 1$), we set $d$ to 4 or 8, depending on the desired compression rate.

We have so far considered networks where each layer has a single parent and a single child
(\ie, the topology is a chain).
We now consider architectures where some layers may have more than one child or more than one parent.

\begin{figure*}[!bt]
	\centering
  \vspace{-1em}
  \includegraphics[width=0.95\linewidth,trim=0mm 144mm 0mm 0mm,clip=true]{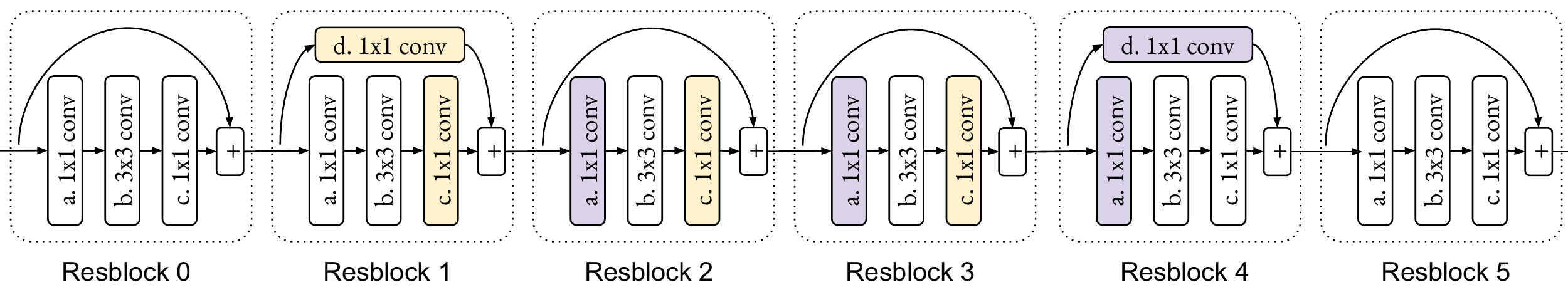}
	\caption{
    \small
    \textbf{Parent-child dependencies in the Resblocks of a ResNet-50 architecture.}
    Purple nodes are children and yellow nodes are parents, and must share the same permutation
    (in $C_\mathrm{in}$ for children and in $C_\mathrm{out}$ for parents) for the network to produce the same output.
  }
  \label{fig:res50}
  \vspace{-2mm}
\end{figure*}

\paragraph{Extension beyond chain architectures:}
\label{sec:permutation_method}
AlexNet~\cite{alexnet} and VGG~\cite{vgg} are examples of popular architectures with a chain topology.
As a consequence, each layer can have a different permutation.
However, architectures with more complicated topologies have more constraints on the permutations that they admit.

For example, consider Figure~\ref{fig:res50}, which depicts six resblocks as used in the popular ResNet-50 architecture.
We start by finding a permutation for layer \texttt{4a}, and realize that its parent is layer \texttt{3c}.
We also notice that
layers \texttt{3c} and \texttt{2c} must share the same permutation for the residual addition to have matching channels.
By induction, this is also true of layers \texttt{1c} and \texttt{1d}, which are now all parents of our initial layer \texttt{4a}.
These parents have children of their own (layers \texttt{2a}, \texttt{3a} and \texttt{4d}),
so these must be counted as siblings of \texttt{4a}, and must share the same permutation as \texttt{4a}.
However, note that all \texttt{b} and \texttt{c} layers are only children, so they can have their own independent permutation.

Operations such as reshaping and concatenation in parent layers may also affect the permutations that a layer can tolerate while preserving functional equivalence.
For example, in the detection head of Mask-RCNN~\cite{maskrcnn}, %
the output (of shape $256\times 7 \times 7$) of a convolutional layer is reshaped (to $12\,544$) before before entering a FC layer.
Moreover, the same tensor is used in another convolutional layer (without reshaping) for mask prediction.
Therefore, the FC layer and the child convolutional layer must share the same permutation.
In this case, the FC layer must keep blocks of $7 \times 7 = 49$ contiguous dimensions together
to respect the channel ordering of its parent (and to match the permutation of its sibling).
Determining the maximum set of independent permutations that an arbitrary network may admit (and finding efficient algorithms to do so)
is a problem  we leave for future work.

\subsection{Learning the Encoding}
Our overarching goal is to learn an encoding of each layer such that the \emph{final} output of the network is preserved.
Towards this goal, we search for a set of codes, codebook, and permutation that minimizes the quantization error $E_t$
of  every layer $t$ of the network:
\begin{align}
\label{eq:qerror}
E_t &= \min_{\vec{P}_t, \vec{B}_t, \mathcal{C}_t} \frac{1}{\hat{m}n} \norm{\widehat{\vec{W}}_t - \vec{P}_t\vec{W}_t}_F^2.
\end{align} 

Our optimization procedure consists of three steps:
\begin{enumerate}[topsep=0.5em, partopsep=0em, itemsep=0em]
\item {\bf Permute:}    We search for a permutation of each layer  that results in subvectors that are easier to quantize.
                        We do this by minimizing the determinant of the covariance of the resulting subvectors.
\item {\bf Quantize:}   We obtain codes and codebooks for each layer by minimizing the difference between the approximated weight and the permuted weight.
\item {\bf Fine-tune:}  Finally, we jointly fine-tune all the codebooks with gradient-based optimization by minimizing the loss
                        function of the original network over the training dataset.
\end{enumerate}
We have found that minimizing the quantization error of the network weights (Eq.~\eqref{eq:qerror})
results in small inaccuracies that accumulate over multiple layers reducing performance;
therefore, it is important to jointly fine-tune the network so that it can recover its original accuracy with gradient descent.
We have also observed that \emph{the quality of the intial reconstruction has a direct impact on the final network accuracy}.
We now describe the three steps in detail.

\subsubsection{Permute}
\label{sec:permutation}
In this step, our goal is to estimate a permutation $\vec{P}_t$ such that the permuted weight matrix $\vec{P}_t\vec{W}_t$
has subvectors $\{ \vec{w}^{\vec{P}_t}_{i,j} \}$ that are easily quantizable.
Intuitively, we want to minimize the spread of the vectors,
as more compact vectors can be expressed more accurately given a fixed number of centroids.
We now formalize this intuition and propose a simple algorithm to find good permutations.

\paragraph{A quantization lower bound:}
We assume that the weight subvectors that form the input to the quantization step 
come from a Gaussian distribution, $ \vec{w}^{\vec{P}_t}_{i,j} \sim \mathcal{N}(\vec{0}, \vec{\Sigma}_t)$, with zero-mean
and covariance $\vec{\Sigma}_t \in \R^{d \times d}$, which is a positive semi-definite matrix.
Thanks to rate distortion theory~\cite{gersho1991vector}, we know that the expected reconstruction error $E_t$ must follow
\begin{equation}
  \label{eq:covdet}
  E_t \ge k^{-\frac{2}{d}} d \matrixdeterminant{\vec{\Sigma}_t}^{\frac{1}{d}};
\end{equation}
in other words, the error is lower-bounded by the determinant of the covariance of the subvectors of $\vec{P}_t\vec{W}_t$.
We assume that we have access to a good minimizer such that, roughly, this bound is equal to the reconstruction
error achieved by our quantization algorithm.
Thus, for a fixed target compression bit-rate, we can focus on finding a permutation $\vec{P}_t$
that minimizes $\matrixdeterminant{\vec{\Sigma}_t}$.

\paragraph{Searching for permutations:}
We make use of Expression~\eqref{eq:covdet} and focus on obtaining a permutation $\vec{P}_t$ that minimizes the determinant
of the covariance of the set $\{ \vec{w}^{\vec{P}_t}_{i,j} \}$. %
We follow an argument similar to that of Ge~\etal~\cite{opq}, and
note that the determinant of any positive semi-definite matrix $\vec{\Sigma}_t \in \R^{d \times d}$,
with elements $\sigma^t_{i,j}$, satisfies Hadamard's inequality:
\begin{equation}
  \matrixdeterminant{\vec{\Sigma}_t} \le \prod_{i=1}^d \sigma^t_{i,i};
  \label{eq:hadamard}
\end{equation}
that is, the determinant of $\vec{\Sigma}_t$ is upper-bounded by the product of its diagonal elements.

Motivated by this inequality, we greedily obtain an initial $\vec{P}_t$ that minimizes the product of
the diagonal elements of $\vec{\Sigma}_t$ by creating $d$ buckets of row indices, each with capacity to hold $\hat{m}=m/d$ elements.
We then compute the variance of each row of $\vec{W}_t$, and greedily assign each row index to the non-full bucket
that results in lowest bucket variance.
Finally, we obtain $\vec{P}_t$ by interlacing rows from the buckets so that rows from the same bucket are placed $d$ rows apart.
$K \times K$ convolutions can be handled similarly, assuming that $\vec{P}_t$ has a block structure, and making use of
the more general Fischer's inequality. Please refer to the
\ifarxiv appendix \else supplementary material \fi
for more details.

There are $\mathcal{O}(m!)$ possible permutations of $\vec{W}_t$, so greedy algorithms are bound to have
limitations on the quality of the solution that they can find.
Thus, we refine our solution via stochastic local search~\cite{hoos2004stochastic}.
Specifically, we iteratively improve the candidate permutation by flipping two dimensions chosen at random, and 
keeping the new permutation if it results in a set of subvectors whose covariance has lower determinant $\matrixdeterminant{\vec{\Sigma}_t}$.
We repeat this procedure for a fixed number of iterations, and return the best permutation obtained.

\subsubsection{Quantize}
\label{sec:src}
In this step, we estimate the codes $\Bt$ and codebook $\Ct$ that approximate the permuted weight $\PWt$.
Given a fixed permutation, this is equivalent to the well-known $k$-means problem.
We use an \emph{annealed} quantization algorithm called SR-C 
originally due to~Zeger~\etal~\cite{zeger1992globally}, and recently adapted by Martinez~\etal~\cite{martinez2018lsqpp} to multi-codebook quantization.
Empirically, SR-C achieves lower quantization error than the vanilla $k$-means algorithm,
and is thus a better minimizer of Expr.~\eqref{eq:covdet}.

\paragraph{A stochastic relaxation of clustering:}
The quantization lower bound from Expression~\eqref{eq:covdet} suggests that the $k$-means algorithm can be annealed by scheduling a perturbation
such that the determinant of the covariance of the  set $\{\wPtij\}$ decreases over time.
Due to Hadamard's inequality (\ie, Expresion~\eqref{eq:hadamard}),
this can be achieved by adding decreasing amounts of noise to $\wPtij$ sampled from a zero-mean Gaussian with diagonal covariance.

Therefore, after randomly initializing the codes, we iteratively update the codebook and codes with
a noisy codebook update (which operates on subvectors with additive diagonalized Gaussian noise),
and a standard $k$-means code update.
We decay the noise according to the schedule $(1-(\tau/I))^\gamma$,
where $\tau$ is the current iteration,
$I$ is the total number of update iterations, and $\gamma$ is a constant. 
We use $\gamma = 0.5$ in all our experiments.
For a detailed description, please refer to Algorithm~\ref{alg:src}.

\vspace{-2mm}
\begin{algorithm}[!ht]
  \caption{SR-C: Stochastic relaxation of $k$-means.}
  \label{alg:src}
  \begin{algorithmic}[1] %
      \Procedure{SR-C}{$\{ \wPtij \}, \vec{\Sigma}_t, k, T, \gamma$}

          \State $\Bt \gets$ \Call{InitializeCodes}{$k$}

          \For{$\tau \gets 1,\dots,T$}
            \LeftComment{\,\,\, \,\,\, \# Add scheduled noise to subvectors}
            \For{$\wPtij \in \{ \wPtij \}$}
              \State $\vec{x}_{i,j} \sim \mathcal{N}(\vec{0}, \mathrm{diag}(\vec{\Sigma}_t))$
              \State $\vec{\hat{w}}^{\vec{P}_t}_{i,j} \gets \wPtij + (\vec{x}_{i,j} \times (1- (\tau/I))^\gamma)$
            \EndFor
            \LeftComment{\# Noisy codebook update}
            \State $\Ct \leftarrow \argmin_\mathcal{C} \sum_{i,j} \lVert \vec{\hat{w}}^{\vec{P}_t}_{i,j} - \vec{c}(b_{i,j}) \rVert_2^2$
            \LeftComment{\# Regular codes update}
            \State $\Bt \leftarrow \argmin_\vec{B} \sum_{i,j} \lVert \vec{w}^{\vec{P}_t}_{i,j} - \vec{c}(b_{i,j}) \rVert_2^2$
          \EndFor
          \State \textbf{return} $\Bt, \Ct$%
      \EndProcedure
  \end{algorithmic}
\end{algorithm}
\vspace{-5mm}

\subsubsection{Fine-tune}
\label{sec:finetune}
Encoding each layer independently causes errors in the activations to accumulate, resulting in degradation of performance.
It is thus important to fine-tune the encoding in order to recover the original accuracy of the network.
In particular, we fix the codes and permutations for the remainder of the procedure.

Let $ \mathcal{L} $ be the \emph{original} loss function of the network (\eg, cross-entropy for classification).
We note that $\mathcal{L}$ is differentiable with respect to each of the learned centroids -- since these are continuous -- so
we use the original training set to fine-tune the centroids with gradient-based learning:
\begin{equation}
    \vec{c}(i) \leftarrow \vec{c}(i) - u \left( \pdv{\mathcal{L}}{\vec{c}(i)}, \theta \right),
    \label{eq:sgd}
\end{equation}
where $u(\cdot, \cdot)$ is an update rule (such as SGD, RMSProp~\cite{tieleman2012lecture} or Adam~\cite{adam})
with hyperparameters $\theta$ (such as learning rate, momentum, and decay rates).

%% file: sections/experiments.tex
\section{Experiments}

We test our method on ResNet~\cite{resnet} architectures for image classification and
Mask R-CNN~\cite{maskrcnn} for object detection and instance segmentation.
We compress standard ResNet-18 and ResNet-50 models that have been pre-trained on ImageNet,
taking the weights directly from the PyTorch model zoo.
We train different networks with ${k \in \{ 256, 512, 1024, 2048 \}}$.
We also clamp the size of the codebook for each layer to $ \min(k, n \times C_{\mathrm{out}} / 4)$.

\begin{table}
    \center
    \small
    \input{tables/block_sizes}
    \vspace{-2mm}
    \caption{\small Subvector sizes and compression regimes.}
    \label{tab:block_sizes}
    \vspace{-4mm}
\end{table}

\paragraph{Small vs. large block sizes:}
To further assess the trade-off between compression and accuracy,
we use two compression regimes. 
In the \emph{large blocks} regime, we use a larger subvector size $d$ for each layer,
which allows the weight matrix to be encoded with fewer codes, and thus leads to higher compression rates.
To describe the subvector sizes we use for each layer, we let $ d_{K} $ denote the subvector
size for a convolutional layer with filters of size $ K \times K $. In the special case when $ K = 1 $,
corresponding to a pointwise convolution, we denote the subvector size by $ d_{\mathrm{pw}} $. Finally,
fully-connected layers have a subvector size of $ d_{\mathrm{fc}} $.
We summarize our subvector sizes for each model and compression regime in Table \ref{tab:block_sizes}.

\paragraph{Bit allocation:}
We compress all the fully-connected and convolutional layers of a network. 
However, following~\cite{bitgoesdown}, we do not compress the first convolutional layer (since it occupies less than 0.05\% of the network size),
the bias of the fully-connected layers, or the batchnorm layers.
While we train with 32-bit floats, we store our final model using 16-bit floats,
which has a negligible impact on validation accuracy (less than 0.02\%).
Finally, we fuse batchnorm layers into two vectors,
which can be done with algebraic manipulation and is
a trick normally used to speed up inference.
Please refer to the
\ifarxiv appendix \else supplementary material \fi
for a detailed breakdown of the bit allocation in our models.

\paragraph{Hyperparameters:}
We use a batch size of 128 for ResNet-18 and a batch size of 64 for ResNet-50. 
For annealed $k$-means, we implement SR-C in the GPU, and run it for 1\,000 iterations.
We fine-tune the codebooks for 9 epochs using Adam~\cite{adam}
with an initial learning rate of $10^{-3}$, which is gradually reduced to $10^{-6}$ using cosine annealing~\cite{cosine}.
Fine-tuning is the most expensive part of this process, and takes around 8 hours both for ResNet-18 (with 1 GPU)
and for ResNet-50 (with 4 GPUs). In the latter case, we scale the learning rate by a factor of 4, following Goyal~\etal~\cite{goyal2017accurate}.
For permutation optimization, we perform 1\,000 local search iterations;
this is done in the CPU in parallel for each independent permutation.
This process takes less than 5 minutes for ResNet-18, and about 10 minutes for ResNet-50 on a 12-core CPU.

\paragraph{Baselines:}
We compare the results of our method against a variety of network compression methods:
Binary Weight Network (BWN)~\cite{rastegari2016xnor}, Trained Ternary
Quantization (TTQ)~\cite{zhu2016trained}, ABC-Net~\cite{abcnet}, LR-Net~\cite{lrnet},
Deep Compression (DC)~\cite{han2015deep}, Hardware-Aware Automated Quantization (HAQ)~\cite{haq},
CLIP-Q~\cite{tungmori2019},
Hessian AWare Quantization of Neural Networks with Mixed Precision (HAWQ)~\cite{hawq}, and HAWQ-V2~\cite{dong2019hawqv2}.
We compare extensively against the recently-proposed Bit Goes Down (BGD) method of \cite{bitgoesdown}
because it is the current state of the art by a large margin.
BGD uses as initialization the method due to Wu~\etal~\cite{wu2016quantized}, and thus subsumes it.
All results presented are taken either from the original papers,
or from two additional surveys~\cite{guo2016survey, cheng2017survey}.

\begin{figure*}[!bt]
    \centering
    \includegraphics[width=0.95\linewidth,trim=0mm 4mm 0mm 3mm,clip=true]{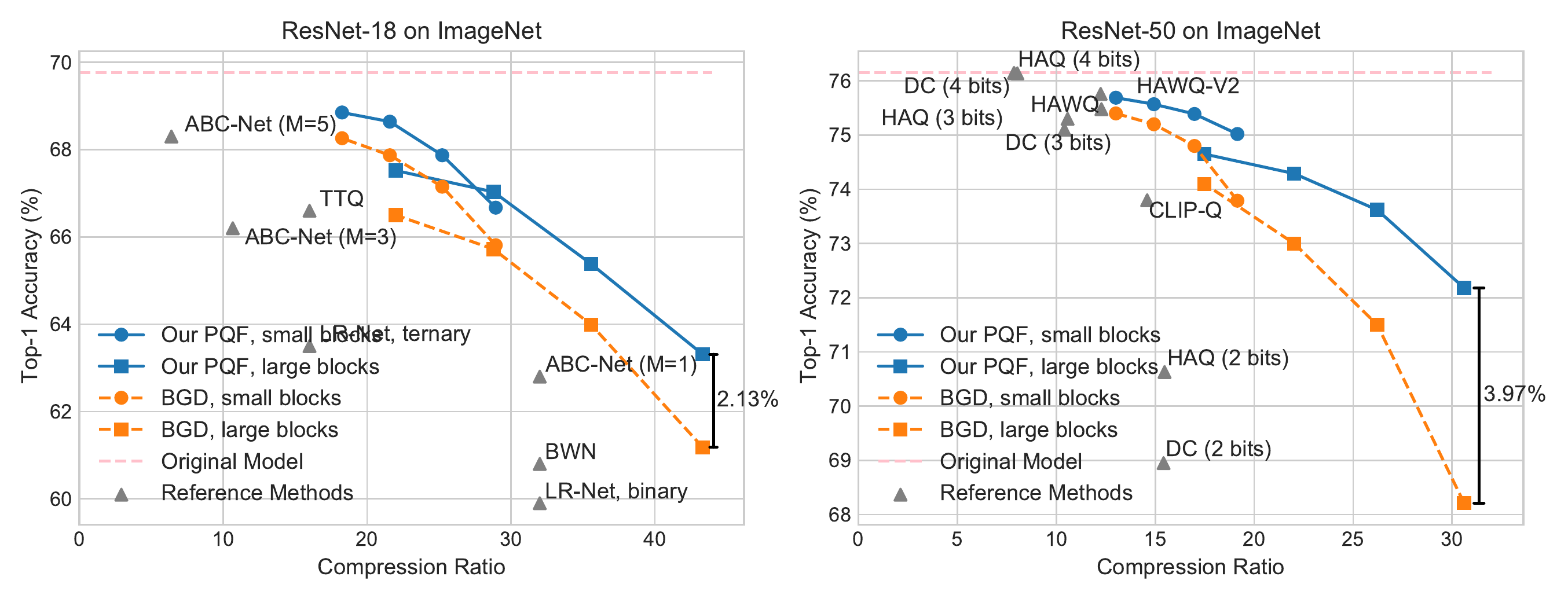}
	\caption{
        \small
		\textbf{Compression results on ResNet-18 and ResNet-50.}
        We compare accuracy vs.\ model size, using models from the PyTorch zoo as a starting point.
        In general, our method achieves higher accuracy compared to previous work.
    }
    \label{fig:samples}
    \vspace{-2mm}
\end{figure*}

\subsection{Image Classification}

A summary of our results can be found in Figure~\ref{fig:samples}.
From the Figure, it is clear that our method outperforms all its competitors.
On ResNet-18 for example, we can surpass the performance of ABC-Net (M=5)
with our \emph{small blocks} models at roughly $3\times$ the compression rate.
Our biggest improvement
generally comes from higher compression rates, and is especially apparent for the larger ResNet-50.
When using large blocks and $k=256$ centroids, we obtain a top-1 accuracy of {\bf 72.18}\% using
only $\sim$3~MB of memory.
This represents an absolute \mytilde{\bf4}\% improvement over the state of the art.
On ResNet-50, our method consistently reduces the remaining error by $40$-$60$\% \wrt{} the state of the art.

\begin{table}
    \small
    \center

\input{tables/semisupervised}
    \vspace{-1mm}
    \caption{
        \small
        {\bf ImageNet classification starting from a semi-supervised ResNet-50.}
        We set a new state of the art in terms of accuracy vs model size.
        $^*$Reproduced from downloaded model.
    }
    \label{tab:semisupervised}
    \vspace{-4mm}
\end{table}

\paragraph{Semi-supervised ResNet-50:}
We also benchmark our method using a stronger backbone as a starting point.
We start from the recently released ResNet-50 model due to Yalniz~\etal~\cite{yalniz2019billion},
which has been pre-trained on unlabelled images from the YFCC100M dataset~\cite{yfcc100m},
and fine-tuned on ImageNet.
While the accuracy of this model is reported to be 79.30\%, we obtain a slightly lower 78.72\%
after downloading the publicly-available model\footnote{\url{https://github.com/facebookresearch/semi-supervised-ImageNet1K-models}};
(contacting the authors we learned that the previous, slightly more accurate model, is no longer available for download).
We use the \emph{small blocks} compression regime with $ k = 256 $,
mirroring the procedure described previously.

We show our results in Table~\ref{tab:semisupervised}, where our model attains a top-1 accuracy of {\bf 77.15}\%.
This means that we are able to outperform previous work by over 1\% absolute accuracy,
with a much smaller gap \wrt the uncompressed model.
We find this result particularly interesting, as we originally expected
distillation to be necessary to transfer the knowledge of the larger
network pretrained on a large corpus of unlabeled images.
However, our results show that at least part of this knowledge is retained through
the initialization and structure that low-error clustering imposes on the compressed network.

\paragraph{Ablation study:}

\begin{table}
    \small
    \center

\input{tables/ablation}
    \vspace{-1mm}
    \caption{
        \small
        {\bf Ablation study.}
        ResNet18 on ImageNet w/large blocks.
    }
    \label{tab:ablation}
    \vspace{-4mm}
\end{table}

In Table~\ref{tab:ablation}, we show results for ResNet-18 using large blocks,
for which we obtain a final accuracy of {\bf 63.31\%}.
We add permutation optimization (Sec.~\ref{sec:permutation}),
annealed $k$-means, as opposed to plain $k$-means (called SR-C in Sec.~\ref{sec:src}),
and the use of the Adam optimizer with cosine annealing instead of plain SGD, as in previous work.
From the Table, we can see that all our components are important and complementary to achieve top accuracy.
It is also interesting to note that a baseline that simply does $k$-means and SGD fine-tuning
is already \mytilde1\% better than the current state-of-the-art.
Since both annealed $k$-means and permutation optimization directly reduce quantization error before fine-tuning,
these experiments demonstrate that minimizing the quantization error of the weights leads to higher final network accuracy.

\subsection{Object Detection and Segmentation}

\begin{table*}[!tb]
    \small
    \centering
    \input{tables/maskrcnn}
    \caption{
        \small
        {\bf Object detection results on MS COCO 2017.}
        We compress a Mask R-CNN network with a ResNet-50 backbone, and include different object detection architectures used by other baselines.
        We report both bounding box (bb) and mask (mk) metrics for Mask R-CNN.
        We also report the accuracy at different IoU when available.
        The memory taken by~\cite{bitgoesdown} corresponds to the (correct) latest version on arXiv.
    }
    \label{tab:maskrcnn}
    \vspace{-3mm}
\end{table*}

\begin{table}[!tb]
    \small
    \centering

\input{tables/maskrcnn_ablation}
    \vspace{-1mm}
    \caption{
        \small
        {\bf Ablation results results on MS COCO 2017.} Permutation optimization is particularly important for Mask-RCNN
    }
    \label{tab:maskrcnn2}
    \vspace{-4mm}
\end{table}

We also benchmark our method on the task of object detection
by compressing the popular ResNet-50 Mask-RCNN FPN architecture~\cite{maskrcnn} using the
MS COCO 2017 dataset~\cite{coco}.
We start from the pretrained model available on the PyTorch model zoo, and apply the same procedure
described above for all the convolutional and linear layers
(plus one deconvolutional layer, which we treat as a convolutional layer for the purpose of compression).
We use the small blocks regime with $k=256$ centroids, for a model of 6.65~MB.

We compress and fine-tune the network on a single Nvidia GTX 1080Ti GPU
with a batch size of 2 for 4 epochs. As before, we use Adam~\cite{adam} and cosine annealing~\cite{cosine},
but with an initial learning rate of $5 \times 10^{-5}$.
Our results are presented in Table~\ref{tab:maskrcnn}.
We also compare against recent baselines such as the Fully Quantized Network (FQN)~\cite{li2019fully},
and the second version of Hessian Aware Quantization (HAWQ-V2)~\cite{dong2019hawqv2},
which showcase results compressing RetinaNet~\cite{retinanet}.

Our method obtains a box AP of 36.3, and a mask AP of 33.5,
which represent improvements of {\bf 2.4}\% and {\bf 2.7}\% over the best previously reported result,
closing the gap to the uncompressed model by 60-70\%.
Compared to BGD~\cite{bitgoesdown}, we also use fewer computational resources,
as they used 8 V100 GPUs and distributed training for compression,
while we use a single 1080Ti GPU.
In Table~\ref{tab:maskrcnn2}, we show again that using both SR-C and permutation optimization is crucial to obtain the best results.
These results demonstrate the ability of our method to generalize to more complex tasks beyond image classification.

%% file: tables/block_sizes.tex
\begin{tabular}{llrrr}
    \toprule
    Model & Regime & $ d_K $ & $ d_{\mathrm{pw}} $ & $ d_{\mathrm{fc}} $\\
    \midrule
    \multirow{2}{*}{\shortstack{ResNet-18}~} & Small blocks~ & $  K^2 $ & 4 & 4 \\
                                             & Large blocks~ & $ 2K^2 $ & 4 & 4 \\
    \midrule
    \multirow{2}{*}{\shortstack{ResNet-50}~} & Small blocks~ & $  K^2 $ & 4 & 4 \\
                                             & Large blocks~ & $ 2K^2 $ & 8 & 4 \\
    \bottomrule
\end{tabular}

%% file: tables/semisupervised.tex
\begin{tabular}{lrrrr}
    \toprule
    & Ratio & Size & Acc. & Gap\\
    \midrule
    Semi-sup R50~\cite{yalniz2019billion} & --~               & 97.50 MB       & 79.30 & --\\
    BGD~\cite{bitgoesdown} &              19$\times$~         & 5.20 MB        & 76.12 & 3.18\\
    \midrule
    Semi-sup R50~\cite{yalniz2019billion} & --~               & 97.50 MB       & $^*$78.72 & -- \\
    Our PQF                               & 19$\times$~       & 5.09 MB        & {\bf 77.15} & 1.57\\
    \bottomrule
\end{tabular}

%% file: tables/ablation.tex
\begin{tabular}{rrrrrr}
  \toprule
  Perm. & SR-C & Adam & Acc. & $\Delta$\\
  \midrule
  &  &  & 62.29 & $-$1.02\\
  \cmark & &  & 62.55 & $-$0.76\\
  \cmark & \cmark & & 62.92 & $-$0.39\\
  \cmark & \cmark & \cmark & {\bf 63.31} & 0.00\\
  \bottomrule
\end{tabular}

%% file: tables/maskrcnn.tex
\begin{tabular}{lrr |rrr |rrr}
    \toprule
    & Size~& Ratio~&
    AP\textsuperscript{bb} & AP$_{50}^{\text{bb}}$ & AP$_{75}^{\text{bb}}$ & AP\textsuperscript{mk} & AP$^{\text{mk}}_{50}$  & AP$^{\text{mk}}_{75}$ \\
    \midrule
    RetinaNet~\cite{retinanet} (uncompressed)           & 145.00 MB &           -- & 35.6 &   -- &   -- &   -- &   -- &   -- \\
    Direct                                              &  18.13 MB &  8.0$\times$ & 31.5 &   -- &   -- &   -- &   -- &   -- \\
    FQN~\cite{li2019fully}                              &  18.13 MB &  8.0$\times$ & 32.5 & 51.5 & 34.7 &   -- &   -- &   -- \\ %
    HAWQ-V2~\cite{dong2019hawqv2}                       &  17.90 MB &  8.1$\times$ & 34.8 &   -- &   -- &   -- &   -- &   -- \\
    \midrule
    \midrule
    Mask-RCNN R-50 FPN~\cite{maskrcnn} (uncompressed)   & 169.40 MB &           -- & 37.9 & 59.2 & 41.1 & 34.6 & 56.0 & 36.8 \\
    BGD~\cite{bitgoesdown}                              &   6.65 MB & 26.0$\times$ & 33.9 & 55.5 & 36.2 & 30.8 & 52.0 & 32.2 \\

    Our PQF                                             &   6.65 MB & 26.0$\times$  & \bf{36.3} & \bf{57.9} & \bf{39.4} & \bf{33.5} & \bf{54.7} & \bf{35.6} \\ 
    \bottomrule
\end{tabular}

%% file: tables/maskrcnn_ablation.tex
\begin{tabular}{lrrrr}
    \toprule
    & Ratio~&
    AP\textsuperscript{bb} & AP\textsuperscript{mk} \\
    \midrule
    Mask-RCNN R-50 FPN~\cite{maskrcnn}   &           -- & 37.9 & 34.6 \\
    BGD~\cite{bitgoesdown}               & 26.0$\times$ & 33.9 & 30.8 \\
    \midrule
    Our PQF (no perm., no SR-C)       & \multirow{3}{*}{\shortstack{26.0$\times$}} & 35.6 & 33.0\\
    Our PQF (no perm.)                &   & 35.8 & 33.1 \\
    Our PQF (full)                        &   & \bf{36.3} & \bf{33.5}\\
    \bottomrule
\end{tabular}

%% file: sections/conclusions.tex
\section{Conclusion}
We have demonstrated that the quantization error of the weights of a neural network is inversely correlated with 
its accuracy after codebook fine tuning.
We have further proposed a method that exploits the functional equivalence of the network under permutation
of its weights to find configurations of the weights that are easier to quantize.
We have also shown that using an annealed $k$-means algorithm further reduces quantization error and improves final network accuracy.
On ResNet-50, our method closes the relative gap to the uncompressed model by 40-70\%
compared to the previous state-of-the-art in a variety of visual tasks.

\ifarxiv
    Our optimization method consists of three stages that focus on different variables of the encoding.
    Future work may focus on techniques that jointly fine-tune the codes and the codebooks,
    or optimization methods that learn the weight permutation jointly with the codes and codebook.
    The determinant of the covariance of the weights is a continuous metric that could be minimized as the
    network is trained from scratch, resulting in networks that are easier to compress by design.
    Last but not least, demonstrating practical hardware acceleration on deep architectures that have been compressed with product codes also remains an open area of research.
\fi

%% file: supp.tex
\section*{A: Permutation initialization for $K \times K$ convolutions}

In Section~\ref{sec:permutation} we describe our goal
of finding a permutation $\vec{P}_t$ of the rows of the weight matrix $\vec{W}_t$,
to create a new matrix $\vec{P}_t\vec{W}_t$ whose subvectors $\{ \vec{w}_{i,j}^{\vec{P}_t} \}$
are easier to quantize.
The algorithm as described works for a linear layer;
that is, when $\vec{W}_t \in \mathbb{R}^{m \times n}$.

We also describe in Section~\ref{sec:permutation_method}
how we reshape the weights of convolutional layers,
$\vec{W} \in \mathbb{R}^{C_\mathrm{in} \times C_\mathrm{out} \times K \times K}$ into a matrix of shape
$\vec{W_r} \in \mathbb{R}^{C_\mathrm{in}K^2 \times C_\mathrm{out}}$ before permutation and quantization.
In the case when $K = 1$ (\ie, pointwise convolutions), we can treat the resulting reshaped weight $\vec{W_r}$
as a linear layer, and apply the same algorithm described in the main paper.
We now show the generalization of our permutation initialization algorithm to $K \times K$ convolutional layers for $K > 1$.

$K \times K$ convolutions have a natural partition for quantization;
namely, clustering the $K^2$ contiguous values that make up a single $K \times K$ convolutional filter.
This natural order has been exploited in previous work~\cite{son2018clustering,bitgoesdown},
and lends itself to acceleration with lookup tables, as commonly done on product-quantized databases~\cite{productquantization}.
Maintaining this partition is also necessary if we want to optimize permutations,
as applying a permutation on the $C_\mathrm{out}$ dimension of a parent layer has the effect of permuting
the order of the channels of the output tensor,
and applying the same permutation on the $C_\mathrm{in}$ dimension of children layers
preserves the function expressed by the network -- this naturally keeps $K \times K$ convolutional filters together.

\begin{figure*}[!bt]
	\centering
  \includegraphics[width=0.9\linewidth,trim=0mm 110mm 0mm 0mm,clip=true]{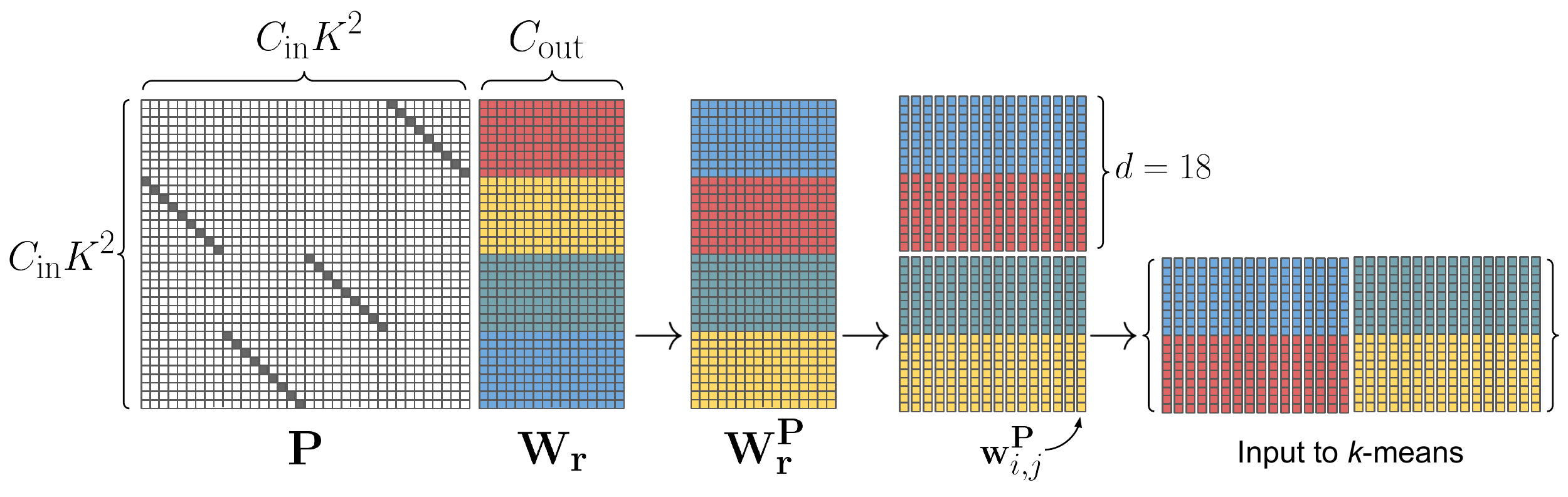}
  \vspace{-0.5em}
	\caption{
    \small
    \textbf{Permutation optimization of a $3 \times 3$ convolutional layer.}
    Our goal is to find a permutation $\vec{P}$ such that the resulting input to $k$-means is easier to quantize.
    We construct $\vec{P}$ with a block structure that swaps $K^2 = 9$ contiguous rows at a time,
    preserving the natural structure of $3 \times 3$ convolutions.
  }
	\label{fig:3cross3}
\end{figure*}

Since the transformed convolution weight matrix
$\vec{W_r} \in \mathbb{R}^{C_\mathrm{in}K^2 \times C_\mathrm{out}}$ contains the $K^2$ values of each filter
stacked across its rows,
we preserve this order by limiting the permutation to only move groups of $K^2$ contiguous rows.
In practice, this forces the permutation matrix $\vec{P}_t$ (of size $C_\mathrm{in}K^2 \times C_\mathrm{in}K^2$)
to have a block structure:
\begin{equation}
  \vec{P}_t = \mqty(
    \vec{\hat{P}}^t_{1,1} & \dots & \vec{\hat{P}}^t_{1,C_\mathrm{in}} \\
    \vdots & \ddots & \vdots \\
    \vec{\hat{P}}^t_{C_\mathrm{in},1} & \dots & \vec{\hat{P}}^t_{C_\mathrm{in},C_\mathrm{in}} \\ 
  ),
\end{equation}
where every submatrix $\vec{\hat{P}}^t_{i,j} \in \{ \vec{0}_{K\times K}, \vec{I}_{K\times K} \}$
is a $K \times K$ matrix with either all zeros, or the identity matrix.
Please refer to Figure~\ref{fig:3cross3} for an illustration of how this matrix permutes blocks of weights.
As before, our goal is to minimize the determinant of the covariance of the resulting subvectors $\{ \vec{w}_{i,j}^{\vec{P}_t} \}$.
We follow Ge~\etal~\cite{opq} and observe that any square, positive semidefinite matrix $\vec{\Sigma}$
may be decomposed into $M^2$ squared blocks $\vec{\hat{\Sigma}}_{i,j}$:
\begin{equation}
  \vec{\Sigma} = \mqty(
    \vec{\hat{\Sigma}}_{1,1} & \dots & \vec{\hat{\Sigma}}_{1,M} \\
    \vdots & \ddots & \vdots \\
    \vec{\hat{\Sigma}}_{M,1} & \dots & \vec{\hat{\Sigma}}_{M,M} \\ 
  ).
\end{equation}
In this case, Fischer's inequality states that
\begin{equation}
  \matrixdeterminant{\vec{\Sigma}} \leq \prod_{i=1}^M \matrixdeterminant{\vec{\hat{\Sigma}}_{i,i}},
\end{equation}
in other words, the determinant of $\vec{\Sigma}$ is bounded by the product of the determinants of its block-diagonal submatrices.
Note that Hadamard's inequality (which we use in our main paper)
can be seen as a corollary of this result.
    To recap, our previously-described algorithm finds an intial permutation that minimizes the product of the diagonal elements by:
    \begin{enumerate}
        \item Creating $d$ buckets
        \item Computing the variance of each row of $\vec{PW_r}$
        \item Assigning each row to the non-full bucket that yields the lowest
              \emph{variance} for the dimensions in that bucket.
    \end{enumerate}
    Thus, we can modify our algorithm to find an initial permutation that minimizes the product
    \emph{of the block-diagonal submatrices} as follows:
    \begin{enumerate}
        \item Instead of creating $d$ buckets, we create $d / K^2$ buckets
        \item Instead of computing the variance of each row of $\vec{PW_r}$, we compute the determinant of the covariance of
        every $K^2$ contiguous rows in $\vec{PW_r}$
        \item We assign each group of $K^2$ contiguous row indices to the non-full bucket that yields the lowest
    \emph{determinant of the covariance} for the dimensions in that bucket.
    \end{enumerate}
Note that for pointwise convolutions,
\ie,~when $K=1$, the above steps naturally yield the algorithm that we described in the main paper.

    After intialization, we also use our iterated local search (ILS) algorithm
    (described in Section~\ref{sec:src})
to find improved permutations. However, instead of swapping pairs of rows of $\vec{PW_r}$, we swap groups of $K^2$ contiguous rows at a time.

Finally, note that this implies that in our current setting we only optimize the permutation of $K \times K$ convolutions
    when the block size is at least $2K^2$.
In practice, this occurs when $d_K = 2K^2 = 2 \cdot 3^2 = 18$ in Table~\ref{tab:block_sizes},
    which corresponds to the large blocks compression regime.
    On the other hand, permutations for $1 \times 1$ convolutions can be optimized under both large and small blocks regimes.

\section*{B: Bit allocation}

Tables~\ref{tab:resnet18}~and~\ref{tab:resnet50} show the bit allocation of ResNet-18 and ResNet-50 under the
small and large compression regimes respectively.
Note that we borrow the bit allocation from~\cite{bitgoesdown}, so our compression ratios match theirs exactly.
As we mention in the main paper, we store the codes using the smallest number of bits possible
\ie, 8 bits for codebooks of size $k=256$, 9 bits for $k=512$, 10 bits for $k=1024$, and 11 bits for $k=2048$.
We also store the codebooks using \texttt{float16},
and layers that are left uncompressed, such as all the batch normalization layers,
are stored with \texttt{float32}.
    We observe that the codes tend to take more memory than the codebooks,
    and that the linear layer tends to be the most memory-heavy part of the network.

\section*{C: Permutation groups}
There are 12 permutation groups for ResNet-18, and 37 groups for ResNet-50.
Listing~\ref{r18_perm} lists the permutations for the default PyTorch implementation of ResNet-18,
and Listings~\ref{r50_perm_1}~and~\ref{r50_perm_2} contain the permutations for ResNet-50. 
These sets of parents and children determine the maximum number of independent permutations that
we can apply without changing the function expressed by the network.
    While we obtained these permutation groups by hand, we believe that computing them automatically on
    arbitrary networks is an interesting area of future work.

\newpage
\begin{footnotesize}
\begin{longtable}{llllr}
  \input{tables/layers_df/resnet18}
  \caption{
    {\bf Bit allocation for Resnet-18.} Small blocks regime, $k=256$.}\\
  \label{tab:resnet18}
\end{longtable}
\end{footnotesize}

\newpage
\begin{footnotesize}
\begin{longtable}{llllr}
  \input{tables/layers_df/resnet50}
  \caption{
    {\bf Bit allocation for Resnet-50.} Large blocks regime, $k=256$.}\\
  \label{tab:resnet50}
\end{longtable}
\end{footnotesize}

\begin{lstfloat}
\begin{python}[ %
  frame=tb,
  caption={
    \small Permutation groups for ResNet-18.
  },
  captionpos=b,
  label=r18_perm,
  backgroundcolor=\color{mygrey}
]

import torchvision.models as models
resnet18 = models.resnet18()

permutations:
    -
      - parents:  [layer1.0.conv1, layer1.0.bn1]
      - children: [layer1.0.conv2]
    -
      - parents:  [layer1.1.conv1, layer1.1.bn1]
      - children: [layer1.1.conv2]
    -
      - parents:  [layer2.0.conv1, layer2.0.bn1]
      - children: [layer2.0.conv2]
    -
      - parents:  [layer2.1.conv1, layer2.1.bn1]
      - children: [layer2.1.conv2]
    -
      - parents:  [layer3.0.conv1, layer3.0.bn1]
      - children: [layer3.0.conv2]
    -
      - parents:  [layer3.1.conv1, layer3.1.bn1]
      - children: [layer3.1.conv2]
    -
      - parents:  [layer4.0.conv1, layer4.0.bn1]
      - children: [layer4.0.conv2]
    -
      - parents:  [layer4.1.conv1, layer4.1.bn1]
      - children: [layer4.1.conv2]
    -
      - parents: [
          conv1, bn1,
          layer1.0.conv2, layer1.0.bn2,
          layer1.1.conv2, layer1.1.bn2,
        ]
      - children: [
          layer1.0.conv1,
          layer2.0.downsample.0,
          layer1.1.conv1,
          layer2.0.conv1,
        ]
    -
      - parents: [
          layer2.0.downsample.0, layer2.0.downsample.1,
          layer2.0.conv2, layer2.0.bn2,
          layer2.1.conv2, layer2.1.bn2,
        ]
      - children: [
          layer2.1.conv1,
          layer3.0.downsample.0,
          layer3.0.conv1,
        ]
    -
      - parents: [
          layer3.0.downsample.0, layer3.0.downsample.1,
          layer3.0.conv2, layer3.0.bn2,
          layer3.1.conv2, layer3.1.bn2,
        ]
      - children: [
          layer3.1.conv1,
          layer4.0.downsample.0,
          layer4.0.conv1,
        ]
    -
      - parents: [
          layer4.0.conv2, layer4.0.bn2,
          layer4.0.downsample.0, layer4.0.downsample.1,
          layer4.1.conv2, layer4.1.bn2,
        ]
      - children: [
          layer4.1.conv1,
          fc,
        ]

\end{python}
\end{lstfloat}

\begin{lstfloat}
  \begin{python}[ %
    frame=tb,
    caption={
      \small Permutation groups for ResNet-50 (part 1/2).
    },
    captionpos=b,
    label=r50_perm_1,
    backgroundcolor=\color{mygrey}
  ]
  
  import torchvision.models as models
  resnet50 = models.resnet50()
  
  permutations:
    -
      - parents: [conv1, bn1]
      - children: [
          layer1.0.conv1,
          layer1.0.downsample.0
        ]
    -
      - parents:  [layer1.0.conv1, layer1.0.bn1]
      - children: [layer1.0.conv2]
    -
      - parents:  [layer1.0.conv2, layer1.0.bn2]
      - children: [layer1.0.conv3]
    -
      - parents: [
        layer1.0.conv3, layer1.0.bn3,
        layer1.0.downsample.0, layer1.0.downsample.1,
        layer1.1.conv3, layer1.1.bn3,
        layer1.2.conv3, layer1.2.bn3,
      ]
      - children: [
          layer1.1.conv1,
          layer1.2.conv1,
          layer2.0.conv1,
          layer2.0.downsample.0
        ]
    -
      - parents:  [layer1.1.conv1, layer1.1.bn1]
      - children: [layer1.1.conv2]
    -
      - parents:  [layer1.1.conv2, layer1.1.bn2]
      - children: [layer1.1.conv3]
    -
      - parents:  [layer1.2.conv1, layer1.2.bn1]
      - children: [layer1.2.conv2]
    -
      - parents:  [layer1.2.conv2, layer1.2.bn2]
      - children: [layer1.2.conv3]
    -
      - parents:  [layer2.0.conv1, layer2.0.bn1]
      - children: [layer2.0.conv2]
    -
      - parents:  [layer2.0.conv2, layer2.0.bn2]
      - children: [layer2.0.conv3]
    -
      - parents: [
          layer2.0.conv3, layer2.0.bn3,
          layer2.0.downsample.0, layer2.0.downsample.1,
          layer2.1.conv3, layer2.1.bn3,
          layer2.2.conv3, layer2.2.bn3,
          layer2.3.conv3, layer2.3.bn3,
        ]
      - children: [
          layer2.1.conv1,
          layer2.2.conv1,
          layer2.3.conv1,
          layer3.0.conv1,
          layer3.0.downsample.0
        ]
    -
      - parents:  [layer2.1.conv1, layer2.1.bn1]
      - children: [layer2.1.conv2]
    -
      - parents:  [layer2.1.conv2, layer2.1.bn2]
      - children: [layer2.1.conv3]
    -
      - parents:  [layer2.2.conv1, layer2.2.bn1]
      - children: [layer2.2.conv2]
    -
      - parents:  [layer2.2.conv2, layer2.2.bn2]
      - children: [layer2.2.conv3]
    -
      - parents:  [layer2.3.conv1, layer2.3.bn1]
      - children: [layer2.3.conv2]
    -
      - parents:  [layer2.3.conv2, layer2.3.bn2]
      - children: [layer2.3.conv3]
    -
      - parents:  [layer3.0.conv1, layer3.0.bn1]
      - children: [layer3.0.conv2]
    -
      - parents:  [layer3.0.conv2, layer3.0.bn2]
      - children: [layer3.0.conv3]

    \end{python}
\end{lstfloat}

\begin{lstfloat}
  \begin{python}[ %
    frame=tb,
    caption={
      \small Permutation groups for ResNet-50 (part 2/2).
    },
    captionpos=b,
    label=r50_perm_2,
    backgroundcolor=\color{mygrey}
  ]
    -
      - parents: [
          layer3.0.conv3, layer3.0.bn3,
          layer3.0.downsample.0, layer3.0.downsample.1,
          layer3.1.conv3, layer3.1.bn3,
          layer3.2.conv3, layer3.2.bn3,
          layer3.3.conv3, layer3.3.bn3,
          layer3.4.conv3, layer3.4.bn3,
          layer3.5.conv3, layer3.5.bn3,
        ]
      - children: [
          layer3.1.conv1,
          layer3.2.conv1,
          layer3.3.conv1,
          layer3.4.conv1,
          layer3.5.conv1,
          layer4.0.conv1,
          layer4.0.downsample.0
        ]
    -
      - parents:  [layer3.1.conv1, layer3.1.bn1]
      - children: [layer3.1.conv2]
    -
      - parents:  [layer3.1.conv2, layer3.1.bn2]
      - children: [layer3.1.conv3]
    -
      - parents:  [layer3.2.conv1, layer3.2.bn1]
      - children: [layer3.2.conv2]
    -
      - parents:  [layer3.2.conv2, layer3.2.bn2]
      - children: [layer3.2.conv3]
    -
      - parents:  [layer3.3.conv1, layer3.3.bn1]
      - children: [layer3.3.conv2]
    -
      - parents:  [layer3.3.conv2, layer3.3.bn2]
      - children: [layer3.3.conv3]
    -
      - parents:  [layer3.4.conv1, layer3.4.bn1]
      - children: [layer3.4.conv2]
    -
      - parents:  [layer3.4.conv2, layer3.4.bn2]
      - children: [layer3.4.conv3]
    -
      - parents:  [layer3.5.conv1, layer3.5.bn1]
      - children: [layer3.5.conv2]
    -
      - parents:  [layer3.5.conv2, layer3.5.bn2]
      - children: [layer3.5.conv3]
    -
      - parents:  [layer4.0.conv1, layer4.0.bn1]
      - children: [layer4.0.conv2]
    -
      - parents:  [layer4.0.conv2, layer4.0.bn2]
      - children: [layer4.0.conv3]
    -
      - parents:  [layer4.1.conv1, layer4.1.bn1]
      - children: [layer4.1.conv2]
    -
      - parents:  [layer4.1.conv2, layer4.1.bn2]
      - children: [layer4.1.conv3]
    -
      - parents:  [layer4.2.conv1, layer4.2.bn1]
      - children: [layer4.2.conv2]
    -
      - parents:  [layer4.2.conv2, layer4.2.bn2]
      - children: [layer4.2.conv3]
    -
      - parents: [
        layer4.0.conv3, layer4.0.bn3,
        layer4.0.downsample.0, layer4.0.downsample.1,
        layer4.1.conv3, layer4.1.bn3,
        layer4.2.conv3, layer4.2.bn3,
      ]
      - children: [
          layer4.1.conv1,
          layer4.2.conv1,
          fc,
        ]
  
  \end{python}
\end{lstfloat}

%% file: tables/layers_df/resnet18.tex
  \\
\toprule
                               {\bf Name} &   {\bf layer\_type} &          {\bf shape} &          {\bf dtype} &     {\bf bits} \\
\midrule
                       conv1.weight &   Conv2d    &  (64, 3, 7, 7) &  torch.float32 &    301056 \\ \midrule
                         bn1.weight &  \multirow{2}{*}{BatchNorm2d} &          (64,) &  torch.float32 &      2048 \\
                           bn1.bias &   &          (64,) &  torch.float32 &      2048 \\\midrule
            layer1.0.conv1.codebook &       \multirow{2}{*}{Conv2d} &       (256, 9) &  torch.float16 &     36864 \\
        layer1.0.conv1.codes\_matrix &        &       (64, 64) &    torch.uint8 &     32768 \\\midrule
                layer1.0.bn1.weight &  \multirow{2}{*}{BatchNorm2d} &          (64,) &  torch.float32 &      2048 \\
                  layer1.0.bn1.bias &   &          (64,) &  torch.float32 &      2048 \\\midrule
            layer1.0.conv2.codebook &       \multirow{2}{*}{Conv2d} &       (256, 9) &  torch.float16 &     36864 \\
        layer1.0.conv2.codes\_matrix &        &       (64, 64) &    torch.uint8 &     32768 \\\midrule
                layer1.0.bn2.weight &  \multirow{2}{*}{BatchNorm2d} &          (64,) &  torch.float32 &      2048 \\
                  layer1.0.bn2.bias &   &          (64,) &  torch.float32 &      2048 \\\midrule
            layer1.1.conv1.codebook &       \multirow{2}{*}{Conv2d} &       (256, 9) &  torch.float16 &     36864 \\
        layer1.1.conv1.codes\_matrix &        &       (64, 64) &    torch.uint8 &     32768 \\\midrule
                layer1.1.bn1.weight &  \multirow{2}{*}{BatchNorm2d} &          (64,) &  torch.float32 &      2048 \\
                  layer1.1.bn1.bias &   &          (64,) &  torch.float32 &      2048 \\\midrule
            layer1.1.conv2.codebook &       \multirow{2}{*}{Conv2d} &       (256, 9) &  torch.float16 &     36864 \\
        layer1.1.conv2.codes\_matrix &        &       (64, 64) &    torch.uint8 &     32768 \\\midrule
                layer1.1.bn2.weight &  \multirow{2}{*}{BatchNorm2d} &          (64,) &  torch.float32 &      2048 \\
                  layer1.1.bn2.bias &   &          (64,) &  torch.float32 &      2048 \\\midrule
            layer2.0.conv1.codebook &       \multirow{2}{*}{Conv2d} &       (256, 9) &  torch.float16 &     36864 \\
        layer2.0.conv1.codes\_matrix &        &      (128, 64) &    torch.uint8 &     65536 \\\midrule
                layer2.0.bn1.weight &  \multirow{2}{*}{BatchNorm2d} &         (128,) &  torch.float32 &      4096 \\
                  layer2.0.bn1.bias &   &         (128,) &  torch.float32 &      4096 \\\midrule
            layer2.0.conv2.codebook &       \multirow{2}{*}{Conv2d} &       (256, 9) &  torch.float16 &     36864 \\
        layer2.0.conv2.codes\_matrix &        &     (128, 128) &    torch.uint8 &    131072 \\\midrule
                layer2.0.bn2.weight &  \multirow{2}{*}{BatchNorm2d} &         (128,) &  torch.float32 &      4096 \\
                  layer2.0.bn2.bias &   &         (128,) &  torch.float32 &      4096 \\\midrule
     layer2.0.downsample.0.codebook &       \multirow{2}{*}{Conv2d} &       (256, 4) &  torch.float16 &     16384 \\
 layer2.0.downsample.0.codes\_matrix &        &      (128, 16) &    torch.uint8 &     16384 \\\midrule
       layer2.0.downsample.1.weight &       \multirow{2}{*}{Conv2d} &         (128,) &  torch.float32 &      4096 \\
         layer2.0.downsample.1.bias &        &         (128,) &  torch.float32 &      4096 \\\midrule
            layer2.1.conv1.codebook &       \multirow{2}{*}{Conv2d} &       (256, 9) &  torch.float16 &     36864 \\
        layer2.1.conv1.codes\_matrix &        &     (128, 128) &    torch.uint8 &    131072 \\\midrule
                layer2.1.bn1.weight &  \multirow{2}{*}{BatchNorm2d} &         (128,) &  torch.float32 &      4096 \\
                  layer2.1.bn1.bias &   &         (128,) &  torch.float32 &      4096 \\\midrule
            layer2.1.conv2.codebook &       \multirow{2}{*}{Conv2d} &       (256, 9) &  torch.float16 &     36864 \\
        layer2.1.conv2.codes\_matrix &        &     (128, 128) &    torch.uint8 &    131072 \\\midrule
                layer2.1.bn2.weight &  \multirow{2}{*}{BatchNorm2d} &         (128,) &  torch.float32 &      4096 \\
                  layer2.1.bn2.bias &   &         (128,) &  torch.float32 &      4096 \\\midrule
            layer3.0.conv1.codebook &       \multirow{2}{*}{Conv2d} &       (256, 9) &  torch.float16 &     36864 \\
        layer3.0.conv1.codes\_matrix &        &     (256, 128) &    torch.uint8 &    262144 \\\midrule
                layer3.0.bn1.weight &  \multirow{2}{*}{BatchNorm2d} &         (256,) &  torch.float32 &      8192 \\
                  layer3.0.bn1.bias &   &         (256,) &  torch.float32 &      8192 \\\midrule
            layer3.0.conv2.codebook &       \multirow{2}{*}{Conv2d} &       (256, 9) &  torch.float16 &     36864 \\
        layer3.0.conv2.codes\_matrix &        &     (256, 256) &    torch.uint8 &    524288 \\\midrule
                layer3.0.bn2.weight &  \multirow{2}{*}{BatchNorm2d} &         (256,) &  torch.float32 &      8192 \\
                  layer3.0.bn2.bias &   &         (256,) &  torch.float32 &      8192 \\\midrule
     layer3.0.downsample.0.codebook &       \multirow{2}{*}{Conv2d} &       (256, 4) &  torch.float16 &     16384 \\
 layer3.0.downsample.0.codes\_matrix &        &      (256, 32) &    torch.uint8 &     65536 \\\midrule
       layer3.0.downsample.1.weight &       \multirow{2}{*}{Conv2d} &         (256,) &  torch.float32 &      8192 \\
         layer3.0.downsample.1.bias &        &         (256,) &  torch.float32 &      8192 \\\midrule
            layer3.1.conv1.codebook &       \multirow{2}{*}{Conv2d} &       (256, 9) &  torch.float16 &     36864 \\
        layer3.1.conv1.codes\_matrix &        &     (256, 256) &    torch.uint8 &    524288 \\\midrule
                layer3.1.bn1.weight &  \multirow{2}{*}{BatchNorm2d} &         (256,) &  torch.float32 &      8192 \\
                  layer3.1.bn1.bias &   &         (256,) &  torch.float32 &      8192 \\\midrule
            layer3.1.conv2.codebook &       \multirow{2}{*}{Conv2d} &       (256, 9) &  torch.float16 &     36864 \\
        layer3.1.conv2.codes\_matrix &        &     (256, 256) &    torch.uint8 &    524288 \\\midrule
                layer3.1.bn2.weight &  \multirow{2}{*}{BatchNorm2d} &         (256,) &  torch.float32 &      8192 \\
                  layer3.1.bn2.bias &   &         (256,) &  torch.float32 &      8192 \\\midrule
            layer4.0.conv1.codebook &       \multirow{2}{*}{Conv2d} &       (256, 9) &  torch.float16 &     36864 \\
        layer4.0.conv1.codes\_matrix &        &     (512, 256) &    torch.uint8 &   1048576 \\\midrule
                layer4.0.bn1.weight &  \multirow{2}{*}{BatchNorm2d} &         (512,) &  torch.float32 &     16384 \\
                  layer4.0.bn1.bias &   &         (512,) &  torch.float32 &     16384 \\\midrule
            layer4.0.conv2.codebook &       \multirow{2}{*}{Conv2d} &       (256, 9) &  torch.float16 &     36864 \\
        layer4.0.conv2.codes\_matrix &        &     (512, 512) &    torch.uint8 &   2097152 \\\midrule
                layer4.0.bn2.weight &  \multirow{2}{*}{BatchNorm2d} &         (512,) &  torch.float32 &     16384 \\
                  layer4.0.bn2.bias &   &         (512,) &  torch.float32 &     16384 \\\midrule
     layer4.0.downsample.0.codebook &       \multirow{2}{*}{Conv2d} &       (256, 4) &  torch.float16 &     16384 \\
 layer4.0.downsample.0.codes\_matrix &        &      (512, 64) &    torch.uint8 &    262144 \\\midrule
       layer4.0.downsample.1.weight &       \multirow{2}{*}{Conv2d} &         (512,) &  torch.float32 &     16384 \\
         layer4.0.downsample.1.bias &        &         (512,) &  torch.float32 &     16384 \\\midrule
            layer4.1.conv1.codebook &       \multirow{2}{*}{Conv2d} &       (256, 9) &  torch.float16 &     36864 \\
        layer4.1.conv1.codes\_matrix &        &     (512, 512) &    torch.uint8 &   2097152 \\\midrule
                layer4.1.bn1.weight &  \multirow{2}{*}{BatchNorm2d} &         (512,) &  torch.float32 &     16384 \\
                  layer4.1.bn1.bias &   &         (512,) &  torch.float32 &     16384 \\\midrule
            layer4.1.conv2.codebook &       \multirow{2}{*}{Conv2d} &       (256, 9) &  torch.float16 &     36864 \\
        layer4.1.conv2.codes\_matrix &        &     (512, 512) &    torch.uint8 &   2097152 \\\midrule
                layer4.1.bn2.weight &  \multirow{2}{*}{BatchNorm2d} &         (512,) &  torch.float32 &     16384 \\
                  layer4.1.bn2.bias &   &         (512,) &  torch.float32 &     16384 \\\midrule
                            fc.bias &       \multirow{2}{*}{Linear} &        (1000,) &  torch.float32 &     32000 \\
                        fc.codebook &        &      (2048, 4) &  torch.float16 &    131072 \\
                    fc.codes\_matrix &        &    (1000, 128) &    torch.int16 &   1408000 \\\midrule
                         total\_bits &              &          &           &  12927232 \\
                         total\_bytes &              &          &           &  1615904 \\
                         total\_KB &              &          &           &  1578.03 \\
                         total\_MB &              &           &           &  1.54 MB \\
\bottomrule \\

%% file: tables/layers_df/resnet50.tex
  \\
\toprule
                               {\bf Name} &   {\bf layer\_type} &          {\bf shape} &          {\bf dtype} &     {\bf bits} \\
\midrule
                       conv1.weight &   Conv2d    &  (64, 3, 7, 7) &  torch.float32 &    301056 \\\midrule
                         bn1.weight &  \multirow{2}{*}{BatchNorm2d} &          (64,) &  torch.float32 &      2048 \\
                           bn1.bias &  &          (64,) &  torch.float32 &      2048 \\\midrule
            layer1.0.conv1.codebook &       \multirow{2}{*}{Conv2d} &       (128, 8) &  torch.float16 &     16384 \\
        layer1.0.conv1.codes\_matrix &       &        (64, 8) &    torch.uint8 &      3584 \\\midrule
                layer1.0.bn1.weight &  \multirow{2}{*}{BatchNorm2d} &          (64,) &  torch.float32 &      2048 \\
                  layer1.0.bn1.bias &  &          (64,) &  torch.float32 &      2048 \\\midrule
            layer1.0.conv2.codebook &       \multirow{2}{*}{Conv2d} &      (256, 18) &  torch.float16 &     73728 \\
        layer1.0.conv2.codes\_matrix &       &       (64, 32) &    torch.uint8 &     16384 \\\midrule
                layer1.0.bn2.weight &  \multirow{2}{*}{BatchNorm2d} &          (64,) &  torch.float32 &      2048 \\
                  layer1.0.bn2.bias &  &          (64,) &  torch.float32 &      2048 \\\midrule
            layer1.0.conv3.codebook &       \multirow{2}{*}{Conv2d} &       (256, 8) &  torch.float16 &     32768 \\
        layer1.0.conv3.codes\_matrix &       &       (256, 8) &    torch.uint8 &     16384 \\\midrule
                layer1.0.bn3.weight &  \multirow{2}{*}{BatchNorm2d} &         (256,) &  torch.float32 &      8192 \\
                  layer1.0.bn3.bias &  &         (256,) &  torch.float32 &      8192 \\\midrule
     layer1.0.downsample.0.codebook &       \multirow{2}{*}{Conv2d} &       (256, 8) &  torch.float16 &     32768 \\
 layer1.0.downsample.0.codes\_matrix &       &       (256, 8) &    torch.uint8 &     16384 \\\midrule
       layer1.0.downsample.1.weight &       \multirow{2}{*}{Conv2d} &         (256,) &  torch.float32 &      8192 \\
         layer1.0.downsample.1.bias &       &         (256,) &  torch.float32 &      8192 \\\midrule
            layer1.1.conv1.codebook &       \multirow{2}{*}{Conv2d} &       (256, 8) &  torch.float16 &     32768 \\
        layer1.1.conv1.codes\_matrix &       &       (64, 32) &    torch.uint8 &     16384 \\\midrule
                layer1.1.bn1.weight &  \multirow{2}{*}{BatchNorm2d} &          (64,) &  torch.float32 &      2048 \\
                  layer1.1.bn1.bias &  &          (64,) &  torch.float32 &      2048 \\\midrule
            layer1.1.conv2.codebook &       \multirow{2}{*}{Conv2d} &      (256, 18) &  torch.float16 &     73728 \\
        layer1.1.conv2.codes\_matrix &       &       (64, 32) &    torch.uint8 &     16384 \\\midrule
                layer1.1.bn2.weight &  \multirow{2}{*}{BatchNorm2d} &          (64,) &  torch.float32 &      2048 \\
                  layer1.1.bn2.bias &  &          (64,) &  torch.float32 &      2048 \\\midrule
            layer1.1.conv3.codebook &       \multirow{2}{*}{Conv2d} &       (256, 8) &  torch.float16 &     32768 \\
        layer1.1.conv3.codes\_matrix &       &       (256, 8) &    torch.uint8 &     16384 \\\midrule
                layer1.1.bn3.weight &  \multirow{2}{*}{BatchNorm2d} &         (256,) &  torch.float32 &      8192 \\
                  layer1.1.bn3.bias &  &         (256,) &  torch.float32 &      8192 \\\midrule
            layer1.2.conv1.codebook &       \multirow{2}{*}{Conv2d} &       (256, 8) &  torch.float16 &     32768 \\
        layer1.2.conv1.codes\_matrix &       &       (64, 32) &    torch.uint8 &     16384 \\\midrule
                layer1.2.bn1.weight &  \multirow{2}{*}{BatchNorm2d} &          (64,) &  torch.float32 &      2048 \\
                  layer1.2.bn1.bias &  &          (64,) &  torch.float32 &      2048 \\\midrule
            layer1.2.conv2.codebook &       \multirow{2}{*}{Conv2d} &      (256, 18) &  torch.float16 &     73728 \\
        layer1.2.conv2.codes\_matrix &       &       (64, 32) &    torch.uint8 &     16384 \\\midrule
                layer1.2.bn2.weight &  \multirow{2}{*}{BatchNorm2d} &          (64,) &  torch.float32 &      2048 \\
                  layer1.2.bn2.bias &  &          (64,) &  torch.float32 &      2048 \\\midrule
            layer1.2.conv3.codebook &       \multirow{2}{*}{Conv2d} &       (256, 8) &  torch.float16 &     32768 \\
        layer1.2.conv3.codes\_matrix &       &       (256, 8) &    torch.uint8 &     16384 \\\midrule
                layer1.2.bn3.weight &  \multirow{2}{*}{BatchNorm2d} &         (256,) &  torch.float32 &      8192 \\
                  layer1.2.bn3.bias &  &         (256,) &  torch.float32 &      8192 \\\midrule
            layer2.0.conv1.codebook &       \multirow{2}{*}{Conv2d} &       (256, 8) &  torch.float16 &     32768 \\
        layer2.0.conv1.codes\_matrix &       &      (128, 32) &    torch.uint8 &     32768 \\\midrule
                layer2.0.bn1.weight &  \multirow{2}{*}{BatchNorm2d} &         (128,) &  torch.float32 &      4096 \\
                  layer2.0.bn1.bias &  &         (128,) &  torch.float32 &      4096 \\\midrule
            layer2.0.conv2.codebook &       \multirow{2}{*}{Conv2d} &      (256, 18) &  torch.float16 &     73728 \\
        layer2.0.conv2.codes\_matrix &       &      (128, 64) &    torch.uint8 &     65536 \\\midrule
                layer2.0.bn2.weight &  \multirow{2}{*}{BatchNorm2d} &         (128,) &  torch.float32 &      4096 \\
                  layer2.0.bn2.bias &  &         (128,) &  torch.float32 &      4096 \\\midrule
            layer2.0.conv3.codebook &       \multirow{2}{*}{Conv2d} &       (256, 8) &  torch.float16 &     32768 \\
        layer2.0.conv3.codes\_matrix &        &      (512, 16) &    torch.uint8 &     65536 \\\midrule
                layer2.0.bn3.weight &  \multirow{2}{*}{BatchNorm2d} &         (512,) &  torch.float32 &     16384 \\
                  layer2.0.bn3.bias &  &         (512,) &  torch.float32 &     16384 \\\midrule
     layer2.0.downsample.0.codebook &       \multirow{2}{*}{Conv2d} &       (256, 8) &  torch.float16 &     32768 \\
 layer2.0.downsample.0.codes\_matrix &        &      (512, 32) &    torch.uint8 &    131072 \\\midrule
       layer2.0.downsample.1.weight &       \multirow{2}{*}{Conv2d} &         (512,) &  torch.float32 &     16384 \\
         layer2.0.downsample.1.bias &        &         (512,) &  torch.float32 &     16384 \\\midrule
            layer2.1.conv1.codebook &        \multirow{2}{*}{Conv2d} &       (256, 8) &  torch.float16 &     32768 \\
        layer2.1.conv1.codes\_matrix &        &      (128, 64) &    torch.uint8 &     65536 \\\midrule
                layer2.1.bn1.weight &  \multirow{2}{*}{BatchNorm2d} &         (128,) &  torch.float32 &      4096 \\
                  layer2.1.bn1.bias &  &         (128,) &  torch.float32 &      4096 \\\midrule
            layer2.1.conv2.codebook &        \multirow{2}{*}{Conv2d} &      (256, 18) &  torch.float16 &     73728 \\
        layer2.1.conv2.codes\_matrix &        &      (128, 64) &    torch.uint8 &     65536 \\\midrule
                layer2.1.bn2.weight &  \multirow{2}{*}{BatchNorm2d} &         (128,) &  torch.float32 &      4096 \\
                  layer2.1.bn2.bias &  &         (128,) &  torch.float32 &      4096 \\\midrule
            layer2.1.conv3.codebook &        \multirow{2}{*}{Conv2d} &       (256, 8) &  torch.float16 &     32768 \\
        layer2.1.conv3.codes\_matrix &        &      (512, 16) &    torch.uint8 &     65536 \\\midrule
                layer2.1.bn3.weight &  \multirow{2}{*}{BatchNorm2d} &         (512,) &  torch.float32 &     16384 \\
                  layer2.1.bn3.bias &  &         (512,) &  torch.float32 &     16384 \\\midrule
            layer2.2.conv1.codebook &        \multirow{2}{*}{Conv2d} &       (256, 8) &  torch.float16 &     32768 \\
        layer2.2.conv1.codes\_matrix &        &      (128, 64) &    torch.uint8 &     65536 \\\midrule
                layer2.2.bn1.weight &  \multirow{2}{*}{BatchNorm2d} &         (128,) &  torch.float32 &      4096 \\
                  layer2.2.bn1.bias &  &         (128,) &  torch.float32 &      4096 \\\midrule
            layer2.2.conv2.codebook &        \multirow{2}{*}{Conv2d} &      (256, 18) &  torch.float16 &     73728 \\
        layer2.2.conv2.codes\_matrix &        &      (128, 64) &    torch.uint8 &     65536 \\\midrule
                layer2.2.bn2.weight &  \multirow{2}{*}{BatchNorm2d} &         (128,) &  torch.float32 &      4096 \\
                  layer2.2.bn2.bias &  &         (128,) &  torch.float32 &      4096 \\\midrule
            layer2.2.conv3.codebook &        \multirow{2}{*}{Conv2d} &       (256, 8) &  torch.float16 &     32768 \\
        layer2.2.conv3.codes\_matrix &        &      (512, 16) &    torch.uint8 &     65536 \\\midrule
                layer2.2.bn3.weight &  \multirow{2}{*}{BatchNorm2d} &         (512,) &  torch.float32 &     16384 \\
                  layer2.2.bn3.bias &  &         (512,) &  torch.float32 &     16384 \\\midrule
            layer2.3.conv1.codebook &        \multirow{2}{*}{Conv2d} &       (256, 8) &  torch.float16 &     32768 \\
        layer2.3.conv1.codes\_matrix &        &      (128, 64) &    torch.uint8 &     65536 \\\midrule
                layer2.3.bn1.weight &  \multirow{2}{*}{BatchNorm2d} &         (128,) &  torch.float32 &      4096 \\
                  layer2.3.bn1.bias &  &         (128,) &  torch.float32 &      4096 \\\midrule
            layer2.3.conv2.codebook &        \multirow{2}{*}{Conv2d} &      (256, 18) &  torch.float16 &     73728 \\
        layer2.3.conv2.codes\_matrix &        &      (128, 64) &    torch.uint8 &     65536 \\\midrule
                layer2.3.bn2.weight &  \multirow{2}{*}{BatchNorm2d} &         (128,) &  torch.float32 &      4096 \\
                  layer2.3.bn2.bias &  &         (128,) &  torch.float32 &      4096 \\\midrule
            layer2.3.conv3.codebook &        \multirow{2}{*}{Conv2d} &       (256, 8) &  torch.float16 &     32768 \\
        layer2.3.conv3.codes\_matrix &        &      (512, 16) &    torch.uint8 &     65536 \\\midrule
                layer2.3.bn3.weight &  \multirow{2}{*}{BatchNorm2d} &         (512,) &  torch.float32 &     16384 \\
                  layer2.3.bn3.bias &  &         (512,) &  torch.float32 &     16384 \\\midrule
            layer3.0.conv1.codebook &        \multirow{2}{*}{Conv2d} &       (256, 8) &  torch.float16 &     32768 \\
        layer3.0.conv1.codes\_matrix &        &      (256, 64) &    torch.uint8 &    131072 \\\midrule
                layer3.0.bn1.weight &  \multirow{2}{*}{BatchNorm2d} &         (256,) &  torch.float32 &      8192 \\
                  layer3.0.bn1.bias &  &         (256,) &  torch.float32 &      8192 \\\midrule
            layer3.0.conv2.codebook &        \multirow{2}{*}{Conv2d} &      (256, 18) &  torch.float16 &     73728 \\
        layer3.0.conv2.codes\_matrix &        &     (256, 128) &    torch.uint8 &    262144 \\\midrule
                layer3.0.bn2.weight &  \multirow{2}{*}{BatchNorm2d} &         (256,) &  torch.float32 &      8192 \\
                  layer3.0.bn2.bias &  &         (256,) &  torch.float32 &      8192 \\\midrule
            layer3.0.conv3.codebook &        \multirow{2}{*}{Conv2d} &       (256, 8) &  torch.float16 &     32768 \\
        layer3.0.conv3.codes\_matrix &        &     (1024, 32) &    torch.uint8 &    262144 \\\midrule
                layer3.0.bn3.weight &  \multirow{2}{*}{BatchNorm2d} &        (1024,) &  torch.float32 &     32768 \\
                  layer3.0.bn3.bias &  &        (1024,) &  torch.float32 &     32768 \\\midrule
     layer3.0.downsample.0.codebook &        \multirow{2}{*}{Conv2d} &       (256, 8) &  torch.float16 &     32768 \\
 layer3.0.downsample.0.codes\_matrix &        &     (1024, 64) &    torch.uint8 &    524288 \\\midrule
       layer3.0.downsample.1.weight &        \multirow{2}{*}{Conv2d} &        (1024,) &  torch.float32 &     32768 \\
         layer3.0.downsample.1.bias &        &        (1024,) &  torch.float32 &     32768 \\\midrule
            layer3.1.conv1.codebook &        \multirow{2}{*}{Conv2d} &       (256, 8) &  torch.float16 &     32768 \\
        layer3.1.conv1.codes\_matrix &        &     (256, 128) &    torch.uint8 &    262144 \\\midrule
                layer3.1.bn1.weight &  \multirow{2}{*}{BatchNorm2d} &         (256,) &  torch.float32 &      8192 \\
                  layer3.1.bn1.bias &  &         (256,) &  torch.float32 &      8192 \\\midrule
            layer3.1.conv2.codebook &        \multirow{2}{*}{Conv2d} &      (256, 18) &  torch.float16 &     73728 \\
        layer3.1.conv2.codes\_matrix &        &     (256, 128) &    torch.uint8 &    262144 \\\midrule
                layer3.1.bn2.weight &  \multirow{2}{*}{BatchNorm2d} &         (256,) &  torch.float32 &      8192 \\
                  layer3.1.bn2.bias &  &         (256,) &  torch.float32 &      8192 \\\midrule
            layer3.1.conv3.codebook &        \multirow{2}{*}{Conv2d} &       (256, 8) &  torch.float16 &     32768 \\
        layer3.1.conv3.codes\_matrix &        &     (1024, 32) &    torch.uint8 &    262144 \\\midrule
                layer3.1.bn3.weight &  \multirow{2}{*}{BatchNorm2d} &        (1024,) &  torch.float32 &     32768 \\
                  layer3.1.bn3.bias &  &        (1024,) &  torch.float32 &     32768 \\\midrule
            layer3.2.conv1.codebook &        \multirow{2}{*}{Conv2d} &       (256, 8) &  torch.float16 &     32768 \\
        layer3.2.conv1.codes\_matrix &        &     (256, 128) &    torch.uint8 &    262144 \\\midrule
                layer3.2.bn1.weight &  \multirow{2}{*}{BatchNorm2d} &         (256,) &  torch.float32 &      8192 \\
                  layer3.2.bn1.bias &  &         (256,) &  torch.float32 &      8192 \\\midrule
            layer3.2.conv2.codebook &        \multirow{2}{*}{Conv2d} &      (256, 18) &  torch.float16 &     73728 \\
        layer3.2.conv2.codes\_matrix &        &     (256, 128) &    torch.uint8 &    262144 \\\midrule
                layer3.2.bn2.weight &  \multirow{2}{*}{BatchNorm2d} &         (256,) &  torch.float32 &      8192 \\
                  layer3.2.bn2.bias &  &         (256,) &  torch.float32 &      8192 \\\midrule
            layer3.2.conv3.codebook &        \multirow{2}{*}{Conv2d} &       (256, 8) &  torch.float16 &     32768 \\
        layer3.2.conv3.codes\_matrix &        &     (1024, 32) &    torch.uint8 &    262144 \\\midrule
                layer3.2.bn3.weight &  \multirow{2}{*}{BatchNorm2d} &        (1024,) &  torch.float32 &     32768 \\
                  layer3.2.bn3.bias &  &        (1024,) &  torch.float32 &     32768 \\\midrule
            layer3.3.conv1.codebook &        \multirow{2}{*}{Conv2d} &       (256, 8) &  torch.float16 &     32768 \\
        layer3.3.conv1.codes\_matrix &        &     (256, 128) &    torch.uint8 &    262144 \\\midrule
                layer3.3.bn1.weight &  \multirow{2}{*}{BatchNorm2d} &         (256,) &  torch.float32 &      8192 \\
                  layer3.3.bn1.bias &  &         (256,) &  torch.float32 &      8192 \\\midrule
            layer3.3.conv2.codebook &        \multirow{2}{*}{Conv2d} &      (256, 18) &  torch.float16 &     73728 \\
        layer3.3.conv2.codes\_matrix &       &     (256, 128) &    torch.uint8 &    262144 \\\midrule
                layer3.3.bn2.weight &  \multirow{2}{*}{BatchNorm2d} &         (256,) &  torch.float32 &      8192 \\
                  layer3.3.bn2.bias &  &         (256,) &  torch.float32 &      8192 \\\midrule
            layer3.3.conv3.codebook &        \multirow{2}{*}{Conv2d} &       (256, 8) &  torch.float16 &     32768 \\
        layer3.3.conv3.codes\_matrix &        &     (1024, 32) &    torch.uint8 &    262144 \\\midrule
                layer3.3.bn3.weight &  \multirow{2}{*}{BatchNorm2d} &        (1024,) &  torch.float32 &     32768 \\
                  layer3.3.bn3.bias &  &        (1024,) &  torch.float32 &     32768 \\\midrule
            layer3.4.conv1.codebook &        \multirow{2}{*}{Conv2d} &       (256, 8) &  torch.float16 &     32768 \\
        layer3.4.conv1.codes\_matrix &        &     (256, 128) &    torch.uint8 &    262144 \\\midrule
                layer3.4.bn1.weight &  \multirow{2}{*}{BatchNorm2d} &         (256,) &  torch.float32 &      8192 \\
                  layer3.4.bn1.bias &  &         (256,) &  torch.float32 &      8192 \\\midrule
            layer3.4.conv2.codebook &        \multirow{2}{*}{Conv2d} &      (256, 18) &  torch.float16 &     73728 \\
        layer3.4.conv2.codes\_matrix &        &     (256, 128) &    torch.uint8 &    262144 \\\midrule
                layer3.4.bn2.weight &  \multirow{2}{*}{BatchNorm2d} &         (256,) &  torch.float32 &      8192 \\
                  layer3.4.bn2.bias &  &         (256,) &  torch.float32 &      8192 \\\midrule
            layer3.4.conv3.codebook &        \multirow{2}{*}{Conv2d} &       (256, 8) &  torch.float16 &     32768 \\
        layer3.4.conv3.codes\_matrix &        &     (1024, 32) &    torch.uint8 &    262144 \\\midrule
                layer3.4.bn3.weight &  \multirow{2}{*}{BatchNorm2d} &        (1024,) &  torch.float32 &     32768 \\
                  layer3.4.bn3.bias &  &        (1024,) &  torch.float32 &     32768 \\\midrule
            layer3.5.conv1.codebook &        \multirow{2}{*}{Conv2d} &       (256, 8) &  torch.float16 &     32768 \\
        layer3.5.conv1.codes\_matrix &        &     (256, 128) &    torch.uint8 &    262144 \\\midrule
                layer3.5.bn1.weight &  \multirow{2}{*}{BatchNorm2d} &         (256,) &  torch.float32 &      8192 \\
                  layer3.5.bn1.bias &  &         (256,) &  torch.float32 &      8192 \\\midrule
            layer3.5.conv2.codebook &        \multirow{2}{*}{Conv2d} &      (256, 18) &  torch.float16 &     73728 \\
        layer3.5.conv2.codes\_matrix &        &     (256, 128) &    torch.uint8 &    262144 \\\midrule
                layer3.5.bn2.weight &  \multirow{2}{*}{BatchNorm2d} &         (256,) &  torch.float32 &      8192 \\
                  layer3.5.bn2.bias &  &         (256,) &  torch.float32 &      8192 \\\midrule
            layer3.5.conv3.codebook &        \multirow{2}{*}{Conv2d} &       (256, 8) &  torch.float16 &     32768 \\
        layer3.5.conv3.codes\_matrix &        &     (1024, 32) &    torch.uint8 &    262144 \\\midrule
                layer3.5.bn3.weight &  \multirow{2}{*}{BatchNorm2d} &        (1024,) &  torch.float32 &     32768 \\
                  layer3.5.bn3.bias &  &        (1024,) &  torch.float32 &     32768 \\\midrule
            layer4.0.conv1.codebook &        \multirow{2}{*}{Conv2d} &       (256, 8) &  torch.float16 &     32768 \\
        layer4.0.conv1.codes\_matrix &        &     (512, 128) &    torch.uint8 &    524288 \\\midrule
                layer4.0.bn1.weight &  \multirow{2}{*}{BatchNorm2d} &         (512,) &  torch.float32 &     16384 \\
                  layer4.0.bn1.bias &  &         (512,) &  torch.float32 &     16384 \\\midrule
            layer4.0.conv2.codebook &        \multirow{2}{*}{Conv2d} &      (256, 18) &  torch.float16 &     73728 \\
        layer4.0.conv2.codes\_matrix &        &     (512, 256) &    torch.uint8 &   1048576 \\\midrule
                layer4.0.bn2.weight &  \multirow{2}{*}{BatchNorm2d} &         (512,) &  torch.float32 &     16384 \\
                  layer4.0.bn2.bias &  &         (512,) &  torch.float32 &     16384 \\\midrule
            layer4.0.conv3.codebook &        \multirow{2}{*}{Conv2d} &       (256, 8) &  torch.float16 &     32768 \\
        layer4.0.conv3.codes\_matrix &        &     (2048, 64) &    torch.uint8 &   1048576 \\\midrule
                layer4.0.bn3.weight &  \multirow{2}{*}{BatchNorm2d} &        (2048,) &  torch.float32 &     65536 \\
                  layer4.0.bn3.bias &  &        (2048,) &  torch.float32 &     65536 \\\midrule
     layer4.0.downsample.0.codebook &        \multirow{2}{*}{Conv2d} &       (256, 8) &  torch.float16 &     32768 \\
 layer4.0.downsample.0.codes\_matrix &        &    (2048, 128) &    torch.uint8 &   2097152 \\\midrule
       layer4.0.downsample.1.weight &        \multirow{2}{*}{Conv2d} &        (2048,) &  torch.float32 &     65536 \\
         layer4.0.downsample.1.bias &        &        (2048,) &  torch.float32 &     65536 \\\midrule
            layer4.1.conv1.codebook &        \multirow{2}{*}{Conv2d} &       (256, 8) &  torch.float16 &     32768 \\
        layer4.1.conv1.codes\_matrix &        &     (512, 256) &    torch.uint8 &   1048576 \\\midrule
                layer4.1.bn1.weight &  \multirow{2}{*}{BatchNorm2d} &         (512,) &  torch.float32 &     16384 \\
                  layer4.1.bn1.bias &  &         (512,) &  torch.float32 &     16384 \\\midrule
            layer4.1.conv2.codebook &        \multirow{2}{*}{Conv2d} &      (256, 18) &  torch.float16 &     73728 \\
        layer4.1.conv2.codes\_matrix &        &     (512, 256) &    torch.uint8 &   1048576 \\\midrule
                layer4.1.bn2.weight &  \multirow{2}{*}{BatchNorm2d} &         (512,) &  torch.float32 &     16384 \\
                  layer4.1.bn2.bias &  &         (512,) &  torch.float32 &     16384 \\\midrule
            layer4.1.conv3.codebook &        \multirow{2}{*}{Conv2d} &       (256, 8) &  torch.float16 &     32768 \\
        layer4.1.conv3.codes\_matrix &       &     (2048, 64) &    torch.uint8 &   1048576 \\\midrule
                layer4.1.bn3.weight &  \multirow{2}{*}{BatchNorm2d} &        (2048,) &  torch.float32 &     65536 \\
                  layer4.1.bn3.bias &  &        (2048,) &  torch.float32 &     65536 \\\midrule
            layer4.2.conv1.codebook &        \multirow{2}{*}{Conv2d} &       (256, 8) &  torch.float16 &     32768 \\
        layer4.2.conv1.codes\_matrix &        &     (512, 256) &    torch.uint8 &   1048576 \\\midrule
                layer4.2.bn1.weight &  \multirow{2}{*}{BatchNorm2d} &         (512,) &  torch.float32 &     16384 \\
                  layer4.2.bn1.bias &  &         (512,) &  torch.float32 &     16384 \\\midrule
            layer4.2.conv2.codebook &        \multirow{2}{*}{Conv2d} &      (256, 18) &  torch.float16 &     73728 \\
        layer4.2.conv2.codes\_matrix &        &     (512, 256) &    torch.uint8 &   1048576 \\\midrule
                layer4.2.bn2.weight &  \multirow{2}{*}{BatchNorm2d} &         (512,) &  torch.float32 &     16384 \\
                  layer4.2.bn2.bias &  &         (512,) &  torch.float32 &     16384 \\\midrule
            layer4.2.conv3.codebook &        \multirow{2}{*}{Conv2d} &       (256, 8) &  torch.float16 &     32768 \\
        layer4.2.conv3.codes\_matrix &        &     (2048, 64) &    torch.uint8 &   1048576 \\\midrule
                layer4.2.bn3.weight &  \multirow{2}{*}{BatchNorm2d} &        (2048,) &  torch.float32 &     65536 \\
                  layer4.2.bn3.bias &  &        (2048,) &  torch.float32 &     65536 \\\midrule
                            fc.bias &       \multirow{2}{*}{Linear} &        (1000,) &  torch.float32 &     32000 \\
                        fc.codebook &        &      (1024, 4) &  torch.float16 &     65536 \\
                    fc.codes\_matrix &       &    (1000, 512) &    torch.int16 &   5120000 \\\midrule
                         total\_bits &              &           &           &  26718976 \\
                         total\_bytes &              &           &           &  3339872  \\
                         total\_KB &              &           &           &  3261.59  \\
                         total\_MB &              &           &           &  3.19 MB \\
\bottomrule \\